\documentclass[lettersize,journal]{IEEEtran}
\usepackage{amsmath,amsfonts}
\usepackage{algorithmic}
\usepackage{algorithm}
\usepackage{array}
\usepackage[caption=false,font=normalsize,labelfont=sf,textfont=sf]{subfig}
\usepackage{textcomp}
\usepackage{stfloats}
\usepackage{url}
\usepackage{verbatim}
\usepackage{graphicx}
\usepackage{cite}
\usepackage{xcolor}
\usepackage{amsthm}
\usepackage{amssymb}
\usepackage{amsmath,lipsum}
\usepackage{epstopdf}
\usepackage{bm}
\usepackage{cuted}
\newcounter{MYtempeqncnt}
\DeclareMathOperator*{\argmin}{argmin}

\begin{document}

\title{Adaptive Decentralized Federated Learning in Energy and Latency Constrained Wireless Networks}

\author{Zhigang Yan and Dong Li,~\IEEEmembership{Senior Member, IEEE}
	\thanks{Zhigang Yan and Dong Li are with the School of Computer
		Science and Engineering, Macau University of Science and Technology, Macau, China. (e-mail: 3220005784@student.must.edu.mo and dli@must.edu.mo).
		
	}
	
}

\maketitle

\begin{abstract}
In Federated Learning (FL), with parameter aggregated by a central node, the communication overhead is a substantial concern. To circumvent this limitation and alleviate the single point of failure within the FL framework, recent studies have introduced Decentralized Federated Learning (DFL) as a viable alternative. Considering the device heterogeneity, and energy cost associated with parameter aggregation, in this paper, the problem on how to efficiently leverage the limited resources available to enhance the model performance is investigated. Specifically, we formulate a problem that minimizes the loss function of DFL while considering energy and latency constraints. The proposed solution involves optimizing the number of local training rounds across diverse devices with varying resource budgets. To make this problem tractable, we first analyze the convergence of DFL with edge devices with different rounds of local training. The derived convergence bound reveals the impact of the rounds of local training on the model performance. Then, based on the derived bound, the closed-form solutions of rounds of local training in different devices are obtained. Meanwhile, since the solutions require the energy cost of aggregation as low as possible, we modify different graph-based aggregation schemes to solve this energy consumption minimization problem, which can be applied to different communication scenarios. Finally, a DFL framework which jointly considers the optimized rounds of local training and the energy-saving aggregation scheme is proposed. Simulation results show that, the proposed algorithm achieves a better performance than the conventional schemes with fixed rounds of local training, and consumes less energy than other traditional aggregation schemes.

\end{abstract}


\section{Introduction}
\IEEEPARstart{I}{n} the present era, the proliferation of Internet of Things (IoT) devices has led to an unprecedented generation and collection of vast volumes of data \cite{yel,yel2}. Simultaneously, the advancement of Artificial Intelligence (AI) research necessitates extensive datasets to train machine learning models. Hence, it becomes imperative to investigate efficient strategies for harnessing these data resources \cite{DL}. However, the transmission of all data to a central node proves impractical due to limitations in communication overhead and privacy concerns \cite{ML}. To mitigate these challenges and facilitate the effective utilization of privacy-sensitive data, the concept of federated learning (FL) emerged \cite{FL1}. FL introduces a distributed learning paradigm that aims to conserve communication resources while enabling collaborative model training on distributed data sources. In FL, devices utilize their available data to train a shared machine learning model by transmitting local model parameters or gradients, thereby avoiding the need to upload all devices' data \cite{FL1,FL2,FL3,FL4}. Nevertheless, FL faces notable hurdles in terms of communication efficiency as the number of participating devices and model parameters continues to grow \cite{com1,com2}. To address this issue, the majority of research efforts have been directed towards parameter compression \cite{q1,q2,q3}, device selection \cite{q3,sch1,sch2,sch3}, and power/bandwidth allocation algorithms \cite{pb1,pb2,pb3} within the FL framework. Despite these endeavors, the conventional FL paradigms continue to rely on a central node for parameter aggregation, leading to heightened communication overhead and vulnerability associated with a single point of failure \cite{c1,c2}. As a means to overcome these limitations and mitigate the risks, recent investigations have explored and introduced the decentralized federated learning (DFL) as a promising alternative approach.

In DFL, parameter aggregation is achieved through direct communication between adjacent devices. Following a sufficient number of communication rounds, the average parameters are synchronized across all participating nodes. Subsequently, each node can iteratively update these aggregated parameters based on its local dataset, until the commencement of the next aggregation round \cite{c2,dfl1}. DFL has demonstrated its versatility and effectiveness as a robust and efficient framework, finding applications in diverse fields such as recommended systems \cite{rec}, vehicle trajectories \cite{v1}, the Industrial IoT \cite{iiot}, and satellite communication \cite{sc}. Moreover, some analyses have been conducted to examine the convergence of DFL, allowing for comprehensive performance evaluations across varying scenarios \cite{conv1,noise1,noise2,error}. These studies have also delved into investigating the impact of communication conditions, including channel fading and noise \cite{noise1,noise2}, and packet errors \cite{error,error2}, on the performance of DFL. 

Despite increasing interest and efforts to address challenges in DFL over wireless networks \cite{ra,error,error2}, a research gap exists, specifically in the realm of device heterogeneity. This heterogeneity, encompassing statistical and system heterogeneity, poses practical issues on the DFL. Statistical heterogeneity involves non-independent and identically distributed (non-i.i.d.) data across devices, potentially affecting the model performance during aggregation. Some aggregation schemes address statistical heterogeneity by selecting neighbors with similar distributions for communication \cite{dfl1,info}. While statistical heterogeneity has been extensively studied, device heterogeneity has received insufficient attention. Variability in device hardware leads to differences in computation and communication capacity and resources, posing inefficiencies when applying the same setting across devices or times. A notable consequence is the straggler effect, causing high latency due to varying computation capacities among devices. An algorithm proposed in \cite{info} addressed this issue by dynamically adjusting the local training frequency. However, beyond variable computation capacities, the DFL system, often implemented over wireless networks, faces heterogeneity in computation and communication resources, including energy costs. This heterogeneity can restrict total local training rounds and updating iterations. However, how to set the limited local training rounds in each iteration to improve the performance has not been paid attention in previous works, since the heterogeneity of resources cost does not appear to have explored this crucial aspect. Considering diverse energy cost budgets among devices, the number of local training rounds in one iteration is influenced not only by convergence time constraints but also by available energy allocation for training. Ensuring more energy for local training requires designing an energy-efficient aggregation scheme to reduce transmission energy costs and consensus errors. 
Unfortunately, only a limited number of works have specifically investigated and addressed the concern about energy consumption on aggregation in the domain of DFL. Notable examples include the application of a quantization scheme based on random linear coding (RLC) to reduce the transmission energy cost \cite{dlc}, the proposal of a balanced communication and computation cost allocation scheme \cite{error2}, the design of an energy-aware algorithm to reduce communication energy costs \cite{sc}, and the development of a bandwidth allocation algorithm to enhance model performance \cite{band}. Despite extensive exploration and discussion of energy budget heterogeneity in the context of FL \cite{pb1,pb2}, these concerns have received limited attention within the realm of DFL. The lack of focused investigation into these issues poses potential detrimental effects to the DFL performance, especially in resource-constrained communication scenarios.

Motivated by these observations, our paper aim to explore the efficient utilization of limited energy resources on each device to enhance the DFL model performance. With fixed total local training rounds due to limited energy budgets, our focus is to design a scheme to allocate these rounds across iterations. To be specific, our goal is to optimize the loss function of the DFL with various rounds of local training on different devices. Therefore, since in the DFL framework where devices have their individual recourse budgets is considered, each device can adjusts its own rounds of local training between two aggregations. To achieve this goal, we formulate a problem to minimize the loss function of DFL by optimizing the rounds of local training in each device, with joint energy and latency cost constraints on each device. To make this problem tractable, we analyze the convergence with the considered DFL framework, which can help us reformulate the original problem as minimizing the derived convergence bound. Furthermore, solving this problem also requires minimizing the energy consumption on the aggregation. Thus, some energy-saving aggregation schemes are applied in our proposed DFL framework, which can reduce the energy cost on the aggregation with different communication conditions (e.g., the channel information between two devices is available or not). Our main contributions are summarized as


\begin{itemize}
	
	\item{We propose a DFL framework with adaptive local training rounds on different iterations and different devices based on their own recourse budgets, and analyze its convergence. Based on the derived convergence bound, we show the impact of rounds of local training on the model performance. Besides, the impact of data distribution on the performance and convergence rate is also reflected.}
	
	\item{According to this framework, we formulate a problem aiming to minimize its loss function with joint energy and latency cost constraints on each device. Using the convergence bound, we relax the original problem as a tractable problem. By solving this problem, the closed-form expressions of the optimal rounds of local training and the energy-saving aggregation scheme to reduce the energy consumption are obtained. Thus, we propose different aggregation schemes based on the Minimum Spanning Tree (MST) algorithm and the Ring-AllReduce algorithm, which are applied to address the aggregation energy cost reduction problem, under different communication conditions (e.g., the channel information is known or not). 
	}
	
	\item{Simulation results not only confirm the theoretical analyses and discussions of the proposed framework, but also make some comparisons with other benchmark schemes. The comparison results show that, the proposed adaptive rounds of local training scheme can achieve a better performance than conventional schemes with fixed rounds, and the proposed aggregation scheme based on the MST or Ring-AllReudce costs less communication energy than the traditional broadcast schemes based on the randomized gossip algorithm.} 
	
\end{itemize}

The remainder of this paper is organized as follows. In Section II, the system model of DFL and its computation and communication are introduced, and the optimization problem is formulated. The convergence analysis and convergence rate of DFL are presented in Section III. We reformulate and decouple the original problem, then obtain its closed-form solutions in Section IV. The simulation results are shown in Section V and conclusions are given in Section VI.

\section{System Model and Problem Formulation}
In this paper, we consider a DFL system consisting of $N$ devices. In DFL, where no central node exists, direct connections can be established between individual nodes. Thus, the DFL system can be represented as an undirected graph denoted as $\mathcal{G}(\mathcal{V},\mathcal{E})$, where $\mathcal{V}={1,2,\cdots,N}$ represents the set of nodes within the graph, and $\mathcal{E}\subseteq\{(i,j)\in\mathcal{V} \times \mathcal{V}~|~i\neq j \}$ represents the set of edges, indicating the connectivity between nodes. Each node has its own dataset, which is written as $\mathcal{D}_{i}$. The whole dataset of DFL is defined by $\mathcal{D}\triangleq\mathcal{D}_{1}\cup\mathcal{D}_{2}\cup\cdots\cup\mathcal{D}_{N}$. 

\subsection{DFL Over Wireless Networks}
We define $\mathbf{w}_{i}\in\mathbb{R}^{k\times 1}$ as the parameter vector of the local model in the $i$-th device. Thus, the parameter vector of the whole DFL system can be written as $\mathbf{w} \triangleq [\mathbf{w}_{1}^{T},\mathbf{w}_{2}^{T},\cdots ,\mathbf{w}_{N}^{T}]^{T}\in\mathbb{R}^{kN\times 1}$ and DFL is done in such a way to find the solution of the optimization problem as follows:
\begin{equation}
	\label{wstar}
	\mathbf{w}^{*} \triangleq \argmin\limits_{\mathbf{w}} F(\mathbf{w},\mathcal{D}),
\end{equation}
where $F(\mathbf{w},\mathcal{D})$ is the global loss function of DFL and it is calculated by
\begin{equation}
	\label{gf}
	F(\mathbf{w},\mathcal{D})= \frac{1}{N}\sum_{i=1}^{N}F_{i}(\mathbf{w}_{i},\mathcal{D}_{i}),
\end{equation}
where $F_{i}(\cdot)$ is the local loss function of the $i$-th device. DFL obtains the value of the global loss function by an iterative algorithm. In the iterative algorithm, this value is obtained through a series of steps.  Each round of iteration involves two fundamental steps: local training and parameter aggregation, as shown in Fig. 1. Specifically, in one iteration, each device updates its local parameter vector via several rounds of local training. Then, all updated parameters are aggregated among the devices and each device begins the next iteration based on the aggregated parameters. Furthermore, it is easy to check that, when all local loss functions are minimized by this iterative algorithm, the global loss function achieve its minimization. Therefore, if we set $\mathbf{w}_{i}^{*}$ as the optimal parameter of $F_{i}(\mathbf{w}_{i})$, we have $\mathbf{w}^{*}=[\mathbf{w}_{1}^{*},\cdots,\mathbf{w}_{N}^{*}]$. Thus, the target of DFL can be written as
\begin{align}
	&\min_{\mathbf{w}_{i}}\quad \frac{1}{N}\sum_{i=1}^{N}F_{i}(\mathbf{w}_{i},\mathcal{D}_{i}) \label{P0}\\
	&\;\textrm{s.t.}\quad \mathbf{w}_{i}=\mathbf{w}_{j},~~~~ \forall i,j. \tag{\ref{P0}{a}} \label{P0a}
\end{align}
Note that, \eqref{P0a} means that, after the aggregations, for all devices, their parameters are the same, which requires the rounds of communication in one round of aggregation is large enough, to guarantee a perfect aggregation. However, in practice, since the resources of communication is limited, the DFL system may not be able to complete enough rounds of communication in one round of aggregation, which causes consensus errors during the aggregation process. This can potentially harm the DFL performance \cite{c2}.

\begin{figure}[!t]
	\centering
	\includegraphics[width=3.2in]{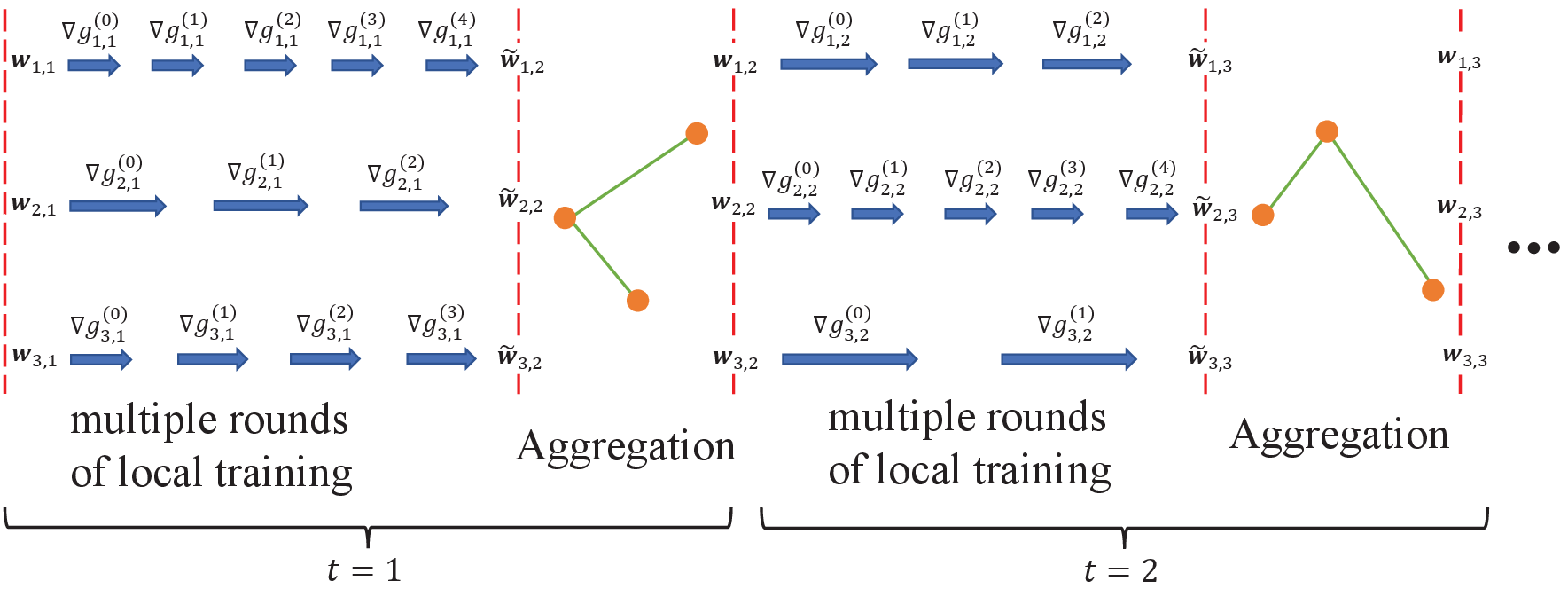}
	\label{fig1}
	\caption{Framework of the iterative algorithm of DFL.}
\end{figure}

During the local training step, each device independently trains local model using its local dataset. This local training process allows devices to update their model parameters by the gradient descent (GD) or stochastic gradient descent (SGD) based on their individual data. In the DFL, since it is usually hard to compute gradients of deep learning models on full datasets, SGD is considered in this work. Thus, the local loss function in \eqref{gf} can be expressed as
\begin{equation}
	\label{sgd0}
	F_{i}(\mathbf{w}_{i},\mathcal{D}_{i})=\mathbb{E}_{\xi_{i}\sim\mathcal{D}_{i}}F_{i}(\mathbf{w}_{i},\xi_{i}),
\end{equation}
where  $\xi_{i}$ is sampled from $\mathcal{D}_{i}$, and the local parameter of the $i$-th device in local training is updated by
\begin{equation}
	\label{sgd}
	\tilde{\mathbf{w}}_{i,t+1} = \mathbf{w}_{i,t}-\eta\sum_{j=0}^{\tau_{i,t}-1}\nabla F_{i}\big(\tilde{\mathbf{w}}_{i,t+\frac{j}{\tau_{i,t}}},\xi_{i}\big),
\end{equation}
where $\tau_{i,t}$ is the number of round of local training in the $t$-th iteration of the $i$-th device, $\tilde{\mathbf{w}}_{i,t+1}$ and $\mathbf{w}_{i,t}$ are the parameter vectors of the $i$-th device after/before the local training in $t$-th rounds of iteration respectively, and $\tilde{\mathbf{w}}_{i,t+\frac{j}{\tau_{i,t}}}$ is the parameter vector of $\tilde{\mathbf{w}}_{i,t}$ after $j$ rounds of local training in $t$-th round of iteration. $\nabla F_{i}(\cdot)\in\mathbb{R}^{k\times 1}$ is the gradient of $\tilde{\mathbf{w}}_{i,t+\frac{j}{\tau_{i,t}}}$, and $\eta$ is the learning rate. Hereinafter, for simplicity, $\mathbf{g}_{i,t}$ and $\mathbf{g}_{i,t}^{(j)}$ are short for $\sum_{j=1}^{\tau_{i,t}}\nabla F_{i}\big(\tilde{\mathbf{w}}_{t+\frac{j}{\tau_{i,t}}},\xi_{i}\big)$ and $\nabla F_{i}\big(\tilde{\mathbf{w}}_{t+\frac{j}{\tau_{i,t}}},\xi_{i}\big)$. Thus, the global parameter vector updated for local training can be written as
\begin{equation}
	\label{sgd2}
	\tilde{\mathbf{w}}_{t+1} = \mathbf{w}_{t}-\eta \mathbf{g}_{t},
\end{equation}
where $\mathbf{g}_{t}=[(\mathbf{g}_{1,t})^{T},\cdots,(\mathbf{g}_{N,t})^{T}]^{T}\in\mathbb{R}^{kN\times 1}$.

Following the local training, the parameter aggregation step takes place. In this step, the updated model parameters from each device are aggregated by communicating with its neighbors. The objective is to synchronize the model parameters across all participating devices. Thus, after the $t$-th round of iteration, the parameter vector of all  devices is
\begin{equation}
	\label{avg}
	\mathbf{w}_{i,t+1} = \frac{1}{N}\sum_{i=1}^{N}\tilde{\mathbf{w}}_{i,t+1},
\end{equation}
and $\mathbf{w}_{t+1} = [\mathbf{w}_{1,t+1}^{T},\mathbf{w}_{2,t+1}^{T},\cdots ,\mathbf{w}_{N,t+1}^{T}]^{T}$, where $\mathbf{w}_{i,t}=\mathbf{w}_{j,t}, \forall i,j,t$. Therefore, the global parameter vector averaged for aggregation can be written as 
\begin{equation}
	\label{avg2}
	\mathbf{w}_{t+1} = \mathbf{P}\tilde{\mathbf{w}}_{t+1},
\end{equation}
where
 \[
\mathbf{P} = \frac{1}{N}\begin{bmatrix}
	\mathbf{I}_{k\times k} & \dots  & \mathbf{I}_{k\times k} \\
	\vdots & \ddots & \vdots \\
	\mathbf{I}_{k\times k}      & \dots   & \mathbf{I}_{k\times k}
\end{bmatrix}_{kN \times kN}  
\]
and $\mathbf{I}$ is the identity matrix. The complete DFL system are shown in Algorithm 1. Indeed, in DFL, the absence of a central node poses a challenge for devices to obtain the aggregated parameters by \eqref{avg} directly. Instead, as mentioned before, devices must engage in communication with their neighbors to averaged the parameters. This process continues until all devices have successfully synchronized the parameters from all participating devices. Thus, modeling the communication between devices is also essential in DFL, which is shown in the following.

\begin{algorithm}[!t]
	\renewcommand{\algorithmicrequire}{\textbf{Input:}}
	\renewcommand{\algorithmicensure}{\textbf{Output:}}
	\caption{Decentralized federated learning (DFL)}
	\label{alg}
	\begin{algorithmic}[1]
		\REQUIRE $\tau_{i,t},\mathbf{w}_{0}$
		\FOR{$t=1,2,\dots, T$}
		\FOR{$j=1,2,\dots, \tau_{i,t}$}
		\STATE For each device $i$ in parallel, its local parameter vector is updated by \eqref{sgd}. 
		\ENDFOR
		\STATE For device $i$, it communicates with its neighbors to average their parameter until all vectors are averaged. 
		\ENDFOR
		\ENSURE  $\mathbf{w}_{T}$
	\end{algorithmic}  
\end{algorithm}

\subsection{Computation and Communication Model}
The resources costs of DFL include the computation and communication costs. In this paper, we consider the latency and energy costs of computation and communication in each device. If the number of processing cycles and the computation capacity (e.g., CPU frequency) of the $i$-th device are represented by $C_{i}$ and $f_{i}$ respectively, the latency and energy cost of one round of local training are given by
\begin{equation}
	\label{ticp}
	L_{i,cp} = C_{i}D_{i}f_{i}^{-1}.
\end{equation}
and
\begin{equation}
	\label{eicp}
	E_{i,cp} = \theta_{i}C_{i}D_{i}f_{i}^{2},
\end{equation}
where $\theta_{i}$ is the effective capacitance coefficient depending on the chipset of the $i$-th device, and $D_{i}$ is the size of dataset $\mathcal{D}_{i}$.

On the other hand, if the $i$-th and $j$-th devices can communicate with each other, we use $h_{ij}$ to denote the channel gain between them, and the communication latency of this communication is given by
\begin{equation}
	\label{ticm}
	L_{ij,cm} = \frac{s}{B\log_{2}\big(1+\frac{p_{t}h_{ij}}{BN_{0}}\big)},
\end{equation}
where $p_{t}$ is the transmission power and $N_{0}$ is the noise power density. The data size of transmitted parameter vector is denoted as $s$. Thus, the energy cost between the $i$-th and $j$-th devices in this communication is given by
\begin{equation}
	\label{eicm}
	E_{ij,cm} = p_{t}L_{ij,cm}=p_{t}\cdot \frac{s}{B\log_{2}\big(1+\frac{p_{t}h_{ij}}{BN_{0}}\big)}.
\end{equation}
In our paper, we model the channel gain $h_{ij}$ as block fading. Thus, we assume that stays invariant within each channel coherence time.

Given the independent energy costs among devices, it is crucial to focus on the individual energy consumption within the DFL framework. Despite this, the collaborative nature of the DFL, where devices collectively train a global model, necessitates the consideration of the entire system's latency. In the DFL system, parameter updates follow a synchronous approach, initiating the global aggregation phase only after all devices complete their local updates. Thus, if the $i$-th device completes $K_{i,t}$ rounds of communication in the $t$-th iteration, the latency of this iteration of DFL is
\begin{equation}
	\label{tcp}
	L_{t,total} = \max\{ \tau_{i,t}L_{i,cp} \}+\max\{K_{i,t}L_{ij,cm}\},
\end{equation}
Besides, the energy cost of one device after $T$ iterations is written as
\begin{equation}
  	\label{ecp}
  	E_{i,total} = \sum_{t=1}^{T}\Big(\tau_{i,t}E_{i,cp}+K_{i,t}\sum_{(i,j)\in\mathcal{E}_{t}}E_{ij,cm}\Big),
\end{equation}
where $\mathcal{E}_{t}$ is the set of the edges, which can represent the connection relationship between devices in $t$-th iteration.
\subsection{Problem Formulation}
To address the need for a cost-efficient DFL system while accounting for device heterogeneity, an optimization problem is formulated. The objective of this optimization problem is to minimize the global loss function by determining the optimal number of rounds of local training for each device and each iteration. Additionally, joint constraints on latency and energy costs are considered. This problem is given by
\begin{align}
	&\min_{\tau_{i,t}}~ F(\mathbf{w}_{T})-F(\mathbf{w}^{*}) \label{P1}\\
	&\;\textrm{s.t.}\quad L_{t,total}\leq \delta_{T},~~~~ \forall t, \tag{\ref{P1}{a}} \label{P1a}\\
	&\quad\quad\; E_{i,total}\leq\delta_{i,E},~~~~ \forall i, \tag{\ref{P1}{b}}\label{P1b}\\
	&\quad\quad\; 1\leq i \leq N,~~~~~~~ i\in\mathbb{N}^{+}, \tag{\ref{P1}{c}}\label{P1c}\\
	&\quad\quad\; 1\leq t \leq T,~~~~~~~\; t\in\mathbb{N}^{+}, \tag{\ref{P1}{d}}\label{P1d}
\end{align}
where $\delta_{T}$ and $\delta_{i,E}$ are the thresholds of the latency and energy cost. The goal of this optimization problem is to achieve a cost-efficient DFL system that maximizes the utilization of available resources while minimizing the global loss function, thereby enhancing the overall performance of DFL. This problem aims to strike a balance between achieving a high model accuracy and optimizing the resource utilization in terms of the energy. By solving this problem, we can obtain a local training scheme with the optimal numbers of local training rounds in each iteration on all devices. This scheme not only consider a better performance of DFL, but also achieve a trade-off between performance and resource costs.
\section{Convergence Analysis of DFL}

In order to tackle \eqref{P1}, a crucial aspect involves analyzing the influence of rounds of local training on the global loss function. Nevertheless, due to the inherent diversity of loss functions across different machine learning models and scenarios, obtaining a general closed-form expression for the loss function about the number of rounds of local training is challenging. To address this issue, we propose an alternative approach by deriving the convergence bound of DFL. Similar to the global loss function, this convergence bound also serves as an evaluation metric to assess the model performance and offers insights into how the number of rounds of local training affects the overall performance of the DFL system.
\subsection{Preliminaries}

To complete the analysis of the convergence of DFL with variable rounds of local training among iterations, we make the following assumptions about the local and global loss functions and their gradients. To distinguish from $\mathbf{g}_{i}$ mentioned before,  $\nabla F_{i}(\mathbf{w}_{i})$ is short for $\nabla F_{i}(\mathbf{w}_{i},\mathcal{D}_{i})$.

\subsubsection*{Assumption 1}

\textit{For any pair of $\mathbf{w}$ and $\mathbf{w}'$, we assume that $F_{i}(\mathbf{w})$ is $m$-strongly convex, $M$-smooth, i.e.,}
\begin{equation}
	\label{c1}
	m\mathbf{I} \preceq \nabla^{2}F_{i}(\mathbf{w}) \preceq M\mathbf{I},~~\forall i, 
\end{equation}
which can also be written as
\begin{equation}
	\begin{aligned}
		\label{c2}
		&F_{i}(\mathbf{w}'){+}\nabla F_{i}(\mathbf{w}')^{T}(\mathbf{w}{-}\mathbf{w}'){+}\frac{m}{2}\Vert \mathbf{w}{-}\mathbf{w}' \Vert_{2}^{2}\leq\\
		&F_{i}(\mathbf{w})\leq F_{i}(\mathbf{w}'){+}\nabla F_{i}(\mathbf{w}')^{T}(\mathbf{w}{-}\mathbf{w}'){+}\frac{M}{2}\Vert \mathbf{w}{-}\mathbf{w}' \Vert_{2}^{2}, 
	\end{aligned}
\end{equation}
where $m$ and $M$ are the minimum and maximum of the singular value of the Hessian matrix of $F_{i}(\mathbf{w})$ (e.g., $\nabla^{2}F_{i}(\mathbf{w})$). This assumption about the loss function are widely adopted in the analysis of FL and DFL, such as \cite{conv1,dlc,ra}.

\textit{Remark 1: According to the expression of $F(\mathbf{w})$ shown in \eqref{gf}, $F(\mathbf{w})$ is also $m$-strongly convex and $M$-smooth.}

\subsubsection*{Assumption 2}

\textit{For any $\mathbf{w}$ and $i$, we respectively define $\sigma$ and $G$ as the upper bounds of the variance of $F_{i}(\mathbf{w}_{i},\xi_{i})$ and its expected value of the $L_{2}$ norm, i.e.,}
\begin{equation}
	\label{g1}
	\mathbb{E}_{\xi_{i}\sim\mathcal{D}_{i}}[\Vert \mathbf{g}_{i} - \nabla F_{i}(\mathbf{w}_{i}) \Vert_{2}^{2}] \leq \sigma,~~\forall i,
\end{equation}
and
\begin{equation}
	\label{g2}
	\mathbb{E}_{\xi_{i}\sim\mathcal{D}_{i}}[\Vert \mathbf{g}_{i} \Vert_{2}^{2}] \leq G,~~\forall i.
\end{equation}

\subsubsection*{Assumption 3}

\textit{We assume that the stochastic gradients $\mathbf{g}_{i}$ are unbiased estimates of the original gradients $\nabla F_{i}(\mathbf{w}_{i})$, which means,}
\begin{equation}
	\label{g3}
	\mathbb{E}_{\xi_{i}\sim\mathcal{D}_{i}}[\mathbf{g}_{i}] = \nabla F_{i}(\mathbf{w}_{i}),~~\forall i.
\end{equation}

These assumptions are also widely applied in the convergence analysis of DFL, such as \cite{dlc,ra}.

\subsubsection*{Assumption 4}

\textit{(Gradient Divergence) for any $\mathbf{w}_{i}$, we assume that}
\begin{equation}
	\label{niid}
	\mathbb{E}_{\xi_{i}\sim\mathcal{D}_{i}}\Big[\Big\Vert \mathbf{g}_{i} -  \frac{1}{N}\sum\limits_{i=1}^{N}\nabla F_{i}(\mathbf{w}_{i}) \Big\Vert_{2}^{2}\Big] \leq \delta_{i}^{2}.
\end{equation}
Then, we define $\delta^{2}\triangleq\frac{1}{N}\sum_{i=1}^{N}\delta_{i}^{2}$ and $\delta_{m}\triangleq\max \delta_{i}^{2}$. This assumption denotes the similarity between the local and global loss functions. Note that, the non-i.i.d. level of the DFL system can be evaluated by $\delta$. When the data distribution in all devices are i.i.d., we have $\delta^{2}=0$. Besides, a larger $\delta^{2}$ means a higher non-i.i.d. level on DFL \cite{sch3,gd1,gd2}.

\subsection{Convergence Bound}
From the definition of global loss function introduced in \eqref{gf}, the global loss function after $T$ iterations is denoted as $F(\mathbf{w}_{T})$. Based on the assumptions and the local training and aggregation approach shown in \eqref{sgd2} and \eqref{avg2}, the gap between the global loss function after the $T$-th iteration and the global optimum is bounded in Theorem 1.

	\textit{Theorem 1: When $\eta\leq\min\{\frac{1}{M},1\}$, the upper bound of the gap between $F(\mathbf{w}_{T})$ and $F(\mathbf{w}^{*})$ is given by}
	\begin{equation}
		\begin{aligned}
			\label{gap1}
			F(\mathbf{w}_{T})&{-}F(\mathbf{w}^{*})\leq\underbrace{\sum\limits_{t=1}^{T}\mathbb{E}[F(\mathbf{w}_{t}){-}F(\tilde{\mathbf{w}}_{t})]}_{\text{Impact of the aggregation}}\\
			&+\underbrace{ \frac{1}{N}\sum\limits_{i=1}^{N}(1{-}m\eta)^{\sum\limits_{t=1}^{T-1}\tau_{i,t}}\big(F_{i}(\mathbf{w}_{i,1}){-}F_{i}(\mathbf{w}_{i}^{*})\big)}_{A_{1}(T)\text{: Impact of the whole local training}}\\
			&+\underbrace{\frac{1}{N}\sum\limits_{i=1}^{N}\frac{\sigma M\eta}{2m}\Big(1+\sum\limits_{j=1}^{T-1}(1{-}m\eta)^{\sum\limits_{t=j}^{T-1}\tau_{i,j}}\Big)}_{A_{2}(t,\tau_{i,t})\text{: Impact of the local training in each iteration}}.
		\end{aligned}
	\end{equation}

\textit{Proof: See Appendix A.$\hfill\qedsymbol$}

\textit{Remark 2: Since the proof of Theorem 1 only relies on the property of $\nabla^{2}F_{i}(\mathbf{w}) \succeq m\mathbf{I}$ under the strong convexity assumption, this convergence result can be extended to a non-convex case. In this case, $F_{i}(\mathbf{w})$ is assumed as an $m'$-bounded non-convexity function (e.g., all the eigenvalues of $\nabla^{2}F_{i}(\mathbf{w})$ are in $[-m',M]$, where $m'<M$). Thus, if we replace the local loss function as a specific regularized version \cite{nonc}, which can make $F_{i}(\mathbf{w})$ and $F(\mathbf{w})$ are still $m'$-strongly convex and $(M+2m')$-Lpischitz. Thus, the proof in Appendix A still applies in this $m'$-bounded non-convexity case.}

From Theorem 1, we know that, the convergence bound of the DFL can be divided into two parts, which are a) Impact of the local training, and b) Impact of the aggregation. Furthermore, from the \eqref{gap1}, the impact of local training consists of two items, i.e., the impact of the whole local training and the impact of the local training in each iteration respectively, which are written as $A_{1}(T)$ and $A_{2}(t,\tau_{i,t})$ in \eqref{gap1}. On one hand, regarding the impact of the whole local training, it can be checked that $A_{1}(T)$ is a decreasing function of $\sum_{t=1}^{T-1}\tau_{i,t}$. Conversely, whatever the specific $\tau_{i,t}$ in the $t$-th iteration is, with the sum of $\tau_{i,t}$ increasing, the gap between $F(\mathbf{w}_{T})$ and $F(\mathbf{w}^{*})$ becomes small. It means that, in the local training phase, increasing the total number of rounds of local training can enhance the model performance. On the other hand, from the expression of $A_{2}(t,\tau_{i,t})$, we can see that, when $\sum_{t=1}^{T-1}\tau_{i,t}$ is fixed, $\sum_{j=1}^{T-1}(1-m\eta)^{\sum_{t=j}^{T-1}\tau_{i,t}}$ becomes lower for small $\tau_{i,t}$. Some further analyses about the impact of specific $\tau_{i,t}$ are given in the Subsection C of the Section IV. Therefore, although a larger total number of local training rounds improves the performance of DFL, the specific $\tau_{i,t}$ allocation scheme in all $T$ iterations can still affect the performance. However, due to the lack of the closed form of the impact of aggregation in the upper bound shown in \eqref{gap1}, it still cannot reflect the convergence of DFL clearly. Thus, to gain further insight into the impact of aggregation, an upper bound of $\sum_{t=1}^{T}\mathbb{E}[F(\mathbf{w}_{t}){-}F(\tilde{\mathbf{w}}_{t})]$ is proposed in the following theorem.

\textit{Theorem 2: When $\eta=\frac{1}{Mt^{4}}$ and $\mathbf{w}_{1}=\mathbf{0}$, the upper bound of $\sum_{t=1}^{T}\mathbb{E}[F(\mathbf{w}_{t}){-}F(\tilde{\mathbf{w}}_{t})]$ is given by}
	\begin{equation}
		\begin{aligned}
			\label{gap2}
			&\sum\limits_{t=1}^{T}\mathbb{E}[F(\mathbf{w}_{t}){-}F(\tilde{\mathbf{w}}_{t})]\leq \frac{\pi^{2}}{12\sqrt{M}}\Bigg(\sqrt{\eta}\Big((\delta_{m}^{2}{+}G)^{2}\\
			&{+}\sigma\big(1{+}(\eta{+}\delta_{m}\sqrt{\eta})M\big)\phi(N)\Big){+}\delta_{m}\phi(N)\Bigg)\triangleq h(\delta_{m},N,t),
		\end{aligned}
	\end{equation}
and
	\begin{equation}
	\begin{aligned}
		\label{gap3}
		\lim\limits_{t\to+\infty}\sum\limits_{t=1}^{T}\mathbb{E}[F(\mathbf{w}_{t})-F(\tilde{\mathbf{w}}_{t})]\leq \frac{\pi^{2}\delta_{m}\phi(N)}{12\sqrt{M}},
	\end{aligned}
\end{equation}
where $\phi(N)\triangleq \frac{k}{N}(N{-}1)(2N{-}1)$ and $k$ is the length of $\mathbf{w}_{i,t}$.

\textit{Proof: See Appendix B.$\hfill\qedsymbol$}

Note that, the accumulation of the impact of aggregation from 1 to $T$-th iteration is determined by the non-i.i.d. level (e.g., $\delta_{m}$) and number of devices (e.g., $N$). From the Assumption 4, we know that, a larger $\delta_{m}$ also means a higher non-i.i.d. level of the DFL system. Thus, $h(\delta_{m},N,t)$, as an increasing function of $\delta_{m}$ and $N$, reveals that, both the higher the non-i.i.d. level and more participating devices lead to the greater the performance damage. In addition, since $\lim\limits_{t\to+\infty}h(\delta_{m},N,t)=\mathcal{O}\big(\delta_{m}\phi(N)\big)$ and $\phi(N)$ is an increasing function with $N$ when $N\geq 1$, in the non-i.i.d. case, a larger $N$ will amplify the impact of $\delta_{m}$ on the performance.

\subsection{Convergence Rate}
In addition to judging whether DFL converges when the number of local training rounds of each device is different, its convergence rate also needs to be paid attention to. When the convergence bound less than $\epsilon$, we consider the DFL is converged. Thus, we can use the minimum number of iterations to achieve this condition to measure the convergence rate of DFL. This value is given by Corollary 1 as follows:

\textit{Corollary 1: Under the condition of convergence, when $\delta_{m}=0$, the convergence rate of DFL is}
\begin{equation}
	\begin{aligned}
		\label{rate}
		T&
		\geq 1+\frac{1}{\epsilon}\Bigg(\frac{\pi^{2}\big(G^{2}+\sigma\phi(N)\big)}{12M}+\frac{\sigma\phi(N)\pi^{2}}{12M}\\
		&\quad\quad\quad
		+\frac{\sigma}{2m}+\frac{F(\mathbf{w}_{1})-F(\mathbf{w}^{*})}{-\ln(1-m\eta)}\Bigg)=\mathcal{O}\Big(\frac{1}{\epsilon}\Big).
	\end{aligned}
\end{equation}

\textit{Proof: See Appendix C.$\hfill\qedsymbol$}

\textit{Remark 3: From Theorem 2, we know that, when $\eta=\frac{1}{Mt^{4}}$ and $\delta_{m}=0$, $\lim\limits_{t\to+\infty}h(\delta_{m},N,t)=0$. Moreover, since $\lim\limits_{t\to+\infty}A_{2}(t,\tau_{i,t})=0$ with $\eta=\frac{1}{Mt^{4}}$ and $\lim\limits_{t\to+\infty}\sum_{t=1}^{T-1}\tau_{i,t}=+\infty$, we have $\lim\limits_{t\to+\infty}h(\delta_{m},N,t)+A_{1}(T)+A_{2}(t,\tau_{i,t})=0$. Thus, the $\epsilon$ in Corollary 1 can be made arbitrarily small.}

Corollary 1 reveals the best case of the convergence rate of DFL. Specifically, when the strongly convex and Lipschitz smooth assumptions hold, in the i.i.d. case, the DFL converges in a linear rate. This result is consistent with previous works on FL and DFL, such as \cite{pb1,dlc,ra}. It means that, although the local training rounds are different among the devices, the convergence of DFL system is the same.

\section{Algorithm Design for Adaptive DFL}

In this section, we reformulate \eqref{P1} by the derived convergence bound and transform it into $N$ set of problems, which include two sub-problems in each set. By solving these two sub-problems and obtaining the closed-form expressions of the solutions, the optimized solutions of the original problem \eqref{P1} and a energy-saving aggregation scheme can be obtained. Finally, the DFL algorithm with adaptive local training among devices is proposed based on these solutions.

\subsection{Problem Reformulation}

Since $\lim\limits_{T\to+\infty}h(\delta_{m},N,T){+}A_{1}(T){+}A_{2}(T,\tau_{i,t})=0$, the upper bound of $F(\mathbf{w}_{T})-F(\mathbf{w}^{*})$ shown in \eqref{gap1} is tight. Thus, the original problem \eqref{P1}, with the goal to minimize $F(\mathbf{w}_{T})-F(\mathbf{w}^{*})$, can be reformulated as minimizing its upper bound, which is
\begin{align}
	&\min_{\tau_{i,t}}\quad h(\delta_{m},N,T)+A_{1}(T)+A_{2}(T,\tau_{i,t})\label{P2}\\
	&\;\textrm{s.t.}\quad \eqref{P1a}-\eqref{P1d}\nonumber,
\end{align}
where the expressions of $h(\delta_{m},N,T)$, $A_{1}(T)$ and $A_{2}(T,\tau_{i,t})$ can be obtained in \eqref{gap1} in Theorem 1.
Note that, in \eqref{P2}, only the values of $A_{1}(T)$ and $A_{2}(T,\tau_{i,t})$ are determined by $\tau_{i,t}$. However, the specific impacts of $\tau_{i,t}$ on $A_{1}(T)$ and $A_{2}(T,\tau_{i,t})$ are different. Specifically, since minimizing $A_{1}(T)$ requires $\sum_{t=1}^{T-1}\tau_{i,t}$ as large as possible with the constraints, the value of each $\tau_{i,t}$ for all $t$ only need to guarantee the sum of $\tau_{i,t}$ is large enough, and the specific value of one $\tau_{i,t}$ is not important. On the contrary, when $\sum_{t=1}^{T-1}\tau_{i,t}$ is fixed, the value of $A_{2}(T,\tau_{i,t})$ is determined on how to allocate these $\tau_{i,t}$ into each iteration. Moreover, the upper bounds of $\sum_{t=1}^{T-1}\tau_{i,t}$ and $\tau_{i,t}$ can be obtained by $\eqref{P1a}$ and $\eqref{P1b}$, which are
\begin{equation}
	\begin{aligned}
		\label{stau}
		\sum\limits_{t=1}^{T-1}\tau_{i,t}\leq \frac{\delta_{i,E}-\sum\limits_{t=1}^{T-1}\Big(K_{i,t}\sum\limits_{(i,j)\in\mathcal{E}_{t}}E_{ij,cm}\Big)}{E_{i,cp}}\triangleq \tau_{\rm total},
	\end{aligned}
\end{equation}
and
\begin{equation}
	\begin{aligned}
		\label{tau}
		\tau_{i,t}\leq \frac{\delta_{T}-K_{i,t}L_{ij,cm}}{L_{i,cp}}\triangleq\tau_{0}.
	\end{aligned}
\end{equation}
Thus, \eqref{P2} can be written as
\begin{align}
	&\min_{\tau_{i,t}}\quad \frac{1}{N}\sum\limits_{i=1}^{N}\Big(\big(F_{i}(\mathbf{w}_{i,1}){-}F_{i}(\mathbf{w}_{i}^{*})\big)(1{-}m\eta)^{\sum\limits_{t=1}^{T-1}\tau_{i,t}}\label{P3}\\
	&~~~~~~~~~~~~~~~~~~~~+\frac{\sigma M\eta}{2m}\sum\limits_{j=1}^{T-1}(1{-}m\eta)^{\sum\limits_{t=j}^{T-1}\tau_{i,j}}\Big)\nonumber\\	
	&\;\textrm{s.t.}\quad \eqref{stau},\eqref{tau},\eqref{P1c},\eqref{P1d}.\nonumber
\end{align}
As analyzed before, solving this problem requires that $\sum_{t=1}^{T-1}\tau_{i,t}$ achieves its maximum. Based on \eqref{stau}, it means the value of $K_{i,t}\sum_{(i,j)\in\mathcal{E}_{t}}E_{ij,cm}$ should be kept as small as possible for all $i$. In addition, since \eqref{P3} is expressed as the weighted average of $N$ items, it can be decoupled into $N$ sets of optimization problems with each set containing two sub-problems, which are
\begin{align}
	&\min_{K_{i,t},\mathcal{E}_{t}}\quad K_{i,t}\sum\limits_{(i,j)\in\mathcal{E}_{t}}E_{ij,cm}\label{P41}\\	
	&\;\textrm{s.t.}\quad\quad\quad \mathcal{E}_{t}\subseteq \mathcal{E}, \tag{\ref{P41}{a}}\label{P41a}
\end{align}
and
\begin{align}
	&\min_{\tau_{i,t}}~\sum\limits_{j=1}^{T-1}(1-m\eta)^{\sum\limits_{t=j}^{T-1}\tau_{i,t}} \label{P42}\\
	&\;\textrm{s.t.}\quad \eqref{tau},\eqref{P1d},\nonumber\\
	&\quad\quad~\sum\limits_{t=1}^{T-1}\tau_{i,t}\leq  \frac{\delta_{i,E}-\sum\limits_{t=1}^{T-1}\Big(K_{i,t}^{*}\sum\limits_{(i,j)\in\mathcal{E}_{t}^{*}}E_{ij,cm}\Big)}{E_{i,cp}},\tag{\ref{P42}{a}}\label{P42a}
\end{align}
where $K_{i,t}^{*}$ and $\mathcal{E}_{t}^{*}$ are the solutions of \eqref{P41}.

\subsection{Proposed Solutions of \eqref{P41}}

Since solving \eqref{P42} needs to obtain $K_{i,t}^{*}$ and $\mathcal{E}_{t}^{*}$, in this subsection, we discuss the solutions of \eqref{P41} firstly. However, note that, the expression of $\mathcal{E}$ is intractable, so it is difficult to solve by some traditional optimization algorithms. In order to circumvent the difficulty in solving \eqref{P41}, we consider two cases, which are channel information between two devices is known or not, and propose different graph-based algorithms to solve it.

\textit{Case 1: Channel information between two devices is known and it stays invariant in one aggregation phase.} 
When the channel information between two devices is known, the transmission energy cost between two devices (e.g., $E_{ij,cm}$) can be calculated by \eqref{eicm}.  Since the connection relationship between devices is modeled as an undirected graph. It is not difficult to find that, if we regard the energy cost of communication between two devices (e.g., $E_{ij,cm}$) as the weight of this edge, minimizing $\sum_{(i,j)\in\mathcal{E}_{t}}E_{ij,cm}$ for all $i$ is equivalent to the problem of finding the connected path in an undirected graph that minimizes the sum of edge weights, which is called Minimum Spanning Tree (MST) generation problem in graph theory.

\begin{algorithm}[!t]
	\renewcommand{\algorithmicrequire}{\textbf{Input:}}
	\renewcommand{\algorithmicensure}{\textbf{Output:}}
	\caption{Kruskal Algorithm to obtain $\mathcal{E}^{*}$ (MST)}
	\label{alg2}
	\begin{algorithmic}[1]
		\REQUIRE The graph modeled by devices $\mathcal{G}(\mathcal{V},\mathcal{E})$. The weights of each edge $E_{ij,cm}$. 
		\STATE Set $\mathcal{E}_{0}=\emptyset$.
		\STATE Sort the elements in $\mathcal{E}$ by their weights into a non-decreasing order, which is written as $\mathcal{E}'$. The element in $\mathcal{E}'$ is denoted as $\mathcal{E}_{ij}'$.
		\STATE Create a disjoint set data structure $\mathcal{S}=\emptyset$.
		\FOR{each $i\in\mathcal{V}$}
		\STATE $\mathcal{S}=\mathcal{S}\cup \{i\}$
		\ENDFOR
		\FOR{each $\mathcal{E}_{ij}'\in\mathcal{E}'$}
		\IF{If no element in $\mathcal{S}$ contains both $i$ and $j$}
		\STATE $\mathcal{E}_{0}=\mathcal{E}_{0}\cup \{i,j\}$
		\STATE Merge all elements of $\mathcal{S}$ containing $i$ or $j$ as a new element and add it to $\mathcal{S}$.
		\ELSE
		\STATE $\mathcal{E}_{0}=\mathcal{E}_{0}$ and $\mathcal{S}=\mathcal{S}$
		\ENDIF
		\STATE $\mathcal{E}_{\rm MST}^{*}=\mathcal{E}_{0}$
		\ENDFOR
		\ENSURE  $\mathcal{E}_{\rm MST}^{*}$
	\end{algorithmic}  
\end{algorithm}

We modify the Kruskal algorithm \cite{mst}, which is one of the classic MST generation algorithms, to achieve the goal of $\sum_{(i,j)\in\mathcal{E}_{t}}E_{ij,cm}$ minimization. Specifically, before each communication period begins, all devices obtain their $E_{ij,cm}$ by \eqref{eicm}, and upload these values to the cloud. After receiving all $E_{ij,cm}$, the cloud finishes Algorithm 2 and outputs the $\mathcal{E}^{*}$. In this algorithm, we create a graph $\mathcal{G}(\mathcal{V},\mathcal{E}_{0})$, where $\mathcal{E}_{0}=\emptyset$. This graph includes all nodes of the original graph $\mathcal{G}(\mathcal{V},\mathcal{E})$, but without any edges. Then, we sort all edges of $\mathcal{G}(\mathcal{V},\mathcal{E})$ by their weights $E_{ij,cm}$ into a non-decreasing order. Based on this order, we put the first element, which is the edge with the minimal weight of $\mathcal{E}_{0}$ and repeat this manipulation on the next elements until all nodes in $\mathcal{G}(\mathcal{V},\mathcal{E}_{0})$ are connected. Then, $\mathcal{E}_{0}$ is transformed into $\mathcal{E}_{\rm MST}^{*}$, which includes a group of edges connecting all nodes, with the minimal sum of weights. It means a least energy cost transmission link in DFL to aggregate all parameters. Finally, devices download $\mathcal{E}_{\rm MST}^{*}$, as the aggregation link with minimal energy costs. Based on this link, one of the devices in the DFL obtain the aggregated parameters. Then, their parameters are transmitted to all others by the same link. This process is summarized as an example in Fig. 2. In this example, the least energy cost transmission link is generated by the Algorithm 2 at first. Then all devices transmit their local parameters over this link to the device 3. Finally, after aggregating all parameters, device 3 transmits them to others through the same link to reach a consensus among all devices. Thus, in this case, $K_{i,t}^{*}=2$.

\begin{figure}[!t]
	\centering
	\includegraphics[width=3.2in]{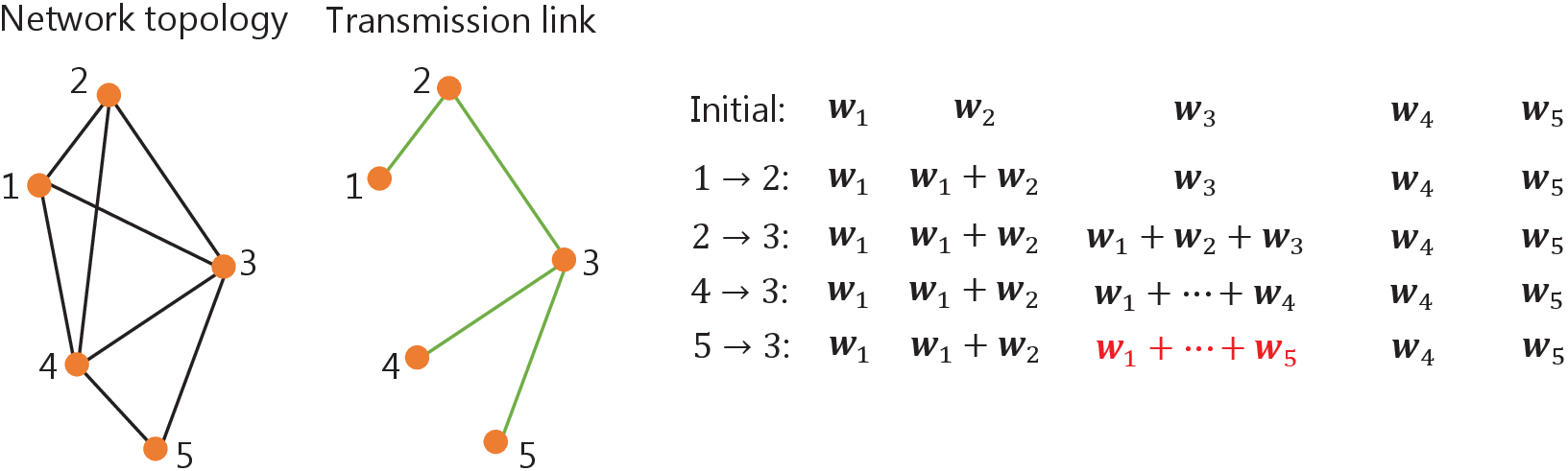}
	\label{fig2}
	\caption{An example of the aggregation scheme based on MST (After the device 3 obtaining the aggregated parameters, it transmits this result to others on the same link).}
\end{figure}

\textit{Case 2: Channel information between two devices is unknown or varying in one aggregation phase.} If we cannot obtain the value of $h_{ij}$ before the aggregation beginning, or the value of $h_{ij}$ cannot keep invariant during the process of the Algorithm 2 working, it is impossible to calculate $\sum_{(i,j)\in\mathcal{E}_{t}}E_{ij,cm}$. Thus, \eqref{P41} can only be minimized by optimizing $K_{i,t}$. The most popular aggregation scheme in the DFL is the randomized gossip algorithm \cite{dfl1}. In this scheme, to guarantee the error caused by the aggregation less than $\varepsilon$, each device need $K_{i,t}=\mathcal{O}(\ln\varepsilon^{-1})$ rounds of communication \cite{gossip}. However, if all devices can connect as a ring topology, another graph-based algorithm, Ring-AllReduce, can be modified to aggregate the parameters. Compared with the randomized gossip algorithm, it can eliminate the consensus error with limited $K_{i,t}$. An example of the Ring-AllReduce algorithm is shown in Fig. 3. In this scheme, all devices transmit their parameters to the next one in parallel. Since the transmission link is a ring, in each round, the parameter of the $i$-th device is sent to the next one for aggregation. Then, the aggregated result is then passed on to the next device in the next round, until it has been traversed across all devices. It is easy to check that, in the Ring-AllReduce algorithm, $K_{i,t}=N-1$. In practice, in addition to the ring topology, nodes in some common topologies can also be connected into a ring, such as the quasi-ring, grid with 2 columns and completed graphs. Thus, in all these topologies, we have $K_{i,t}^{*}=N-1$ and $\mathcal{E}_{t}^{*}=\mathcal{E}_{\rm ring}$.
\begin{figure}[!t]
	\centering
	\includegraphics[width=3.2in]{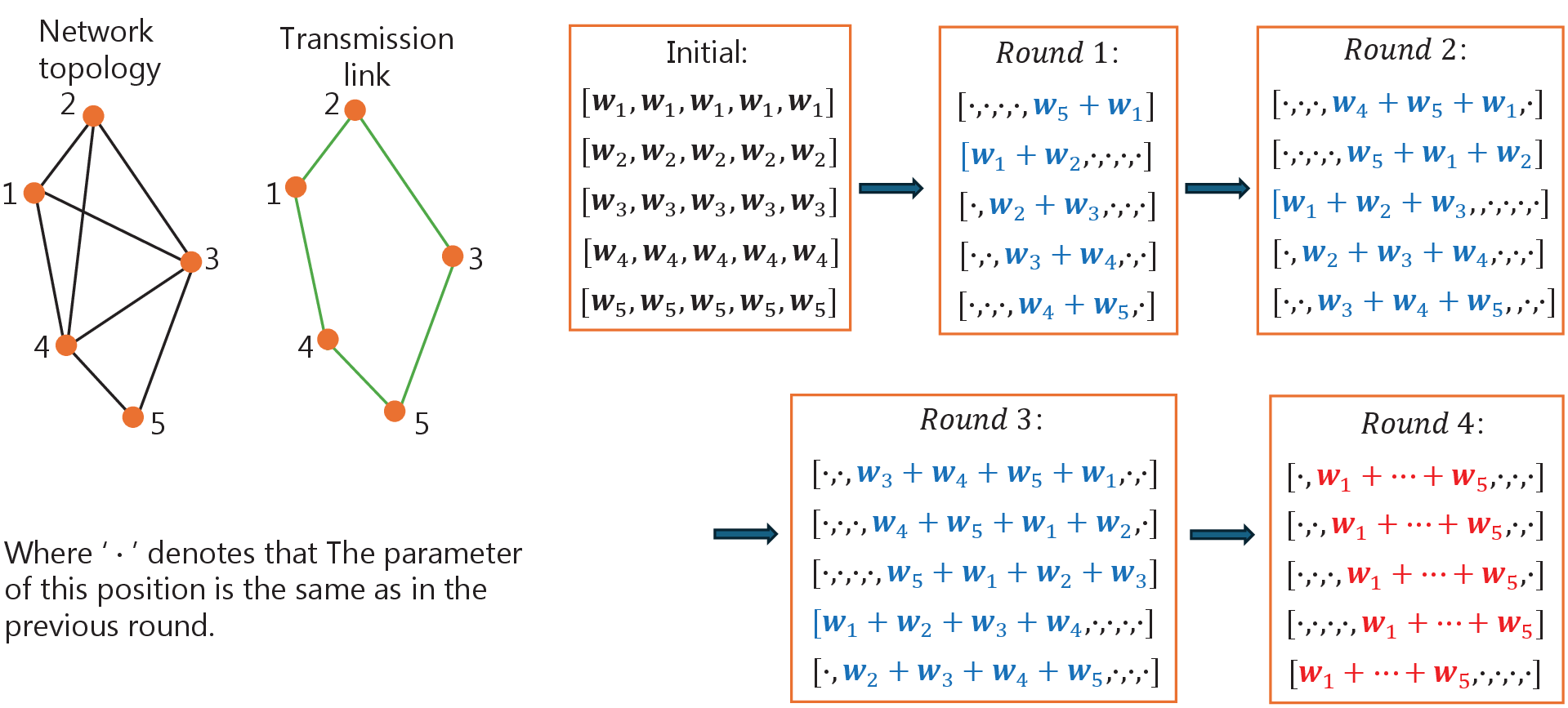}
	\label{fig3}
	\caption{An example of the aggregation scheme based on Ring-AllReduce.}
\end{figure}

Through discussions on the above different cases, the solutions of \eqref{P41} proposed by the mentioned algorithms can be summarized in the following theorem.

\textit{Theorem 3: The solutions of \eqref{P41} in different cases are given by}
\begin{equation}
	\begin{aligned}
		\label{s1}
		(K_{i,t}^{*},\mathcal{E}_{t}^{*})=
		\begin{cases}
			(2,\mathcal{E}_{\rm MST}^{*})~~~~~~~~~~~\!\!\textrm{Case 1},\\
			(N-1,\mathcal{E}_{\rm ring})~~~~~\textrm{Case 2 with ring},\\
			(\mathcal{O}(\ln\varepsilon^{-1}),\mathcal{E})~~~~~\!\textrm{otherwise},
		\end{cases}
	\end{aligned}
\end{equation}
where $\varepsilon$ is the error caused by the aggregation. In Case 1 and Case 2 with ring topology, we have $\varepsilon=0$. 

\textit{Remark 4: Since $\varepsilon$ cannot be completely eliminated in the randomized gossip algorithm, if the usage conditions of MST or Ring-AllReduce are not satisfied, DFL will face a trade-off in the number of communications (e.g., $K_{i,t}$) and consensus error (e.g., $\varepsilon$). The issue about how to set a balanced $K_{i,t}$ to enhance the performance of DFL has been studied in \cite{error2}.}

With the definition of $\mathcal{E}_{\rm MST}^{*}$, we obtain the following corollary, which states the comparison between the solutions of Case 1 and Case 2 with a ring topology.

\textit{Corollary 2: The aggregation scheme based on MST consumes less energy than the scheme based on Ring-AllReduce, which can be expressed as}
	\begin{equation}
			\label{com}
			\begin{aligned}
					2\sum_{(i,j)\in\mathcal{E}_{\rm MST}^{*}}E_{ij,cm}\leq (N-1)\sum_{(i,j)\in\mathcal{E}_{\rm ring}}E_{ij,cm}.
			\end{aligned}
	\end{equation}

\textit{Proof: Similar $\mathcal{E}_{\rm MST}^{*}$, $\mathcal{E}_{\rm ring}$ is also a sub-graph of $\mathcal{E}$. Thus, the sum of the weights of $\mathcal{E}_{\rm ring}$ edges must be larger than $\mathcal{E}_{\rm MST}^{*}$.$\hfill\qedsymbol$}

Corollary 2 shows that, if the channel information is known and it can stay invariant until all devices aggregate the parameters, the aggregation scheme proposed in Case 1 can guarantee the value of the objective function in \eqref{P41} less than modifying the scheme proposed in Case 2. However, when the scenario is more general, such as unknown channel information or time-varying channel, considering the Case 2 is also necessary. In addition, as introduced before, the aggregation scheme in Case 1 is a sequential aggregation approach, thereby facilitating a potentially slower consensus time compared with the aggregation scheme based on Ring-AllReduce or randomized gossip algorithm.

\subsection{Proposed Solutions of \eqref{P42}}

Note that, problem \eqref{P42} is still difficult to solve, since the coupled $\tau_{i,t}$ in \eqref{P42a} leads to the non-convexity of \eqref{P42}. Thus, we give a numerical example to observe the distribution of $\tau_{i,t}$ from $t=1$ to $T-1$ at first. If we set $T=4$ and $\sum_{t=1}^{T-1}\tau_{i,t}=5$, and let $1-m\eta=a$, all possible values of the objective function of \eqref{P42} are shown in Table I. From this result, we can see that, if the DFL system want to achieve a better performance, it prefer to allocate less rounds of local training at the beginning. Specifically, if $\tau_{\rm total}^{*}$ denotes the $\tau_{\rm total}$ with $K_{i,t}^{*}$ and $\mathcal{E}_{t}^{*}$, which can be calculated by \eqref{stau}, minimizing the objective function of \eqref{P42} without any constraints means $\tau_{i,t}=\cdots=\tau_{i,T-2}=1$ and $\tau_{i,T-1}=\lfloor\tau_{\rm total}^{*}\rfloor-T+2$. However, by jointly considering the constraints \eqref{tau} and \eqref{P42a}, obtaining the closed-form solutions of problem \eqref{P42} requires further relaxation of it. Thus, we assume that there exits an auxiliary variable $\zeta$, which satisfies $\sum_{j=1}^{T-1}(1-m\eta)^{\sum_{t=j}^{T-1}\tau_{i,t}}\leq \zeta$. Then, we can relax \eqref{P42} into a problem of minimizing $\zeta$, which is shown as

\begin{align}
	&\min_{\tau_{i,t}}\quad \zeta\label{P5}\\
	&\;\textrm{s.t.}\quad \eqref{tau},\eqref{P41a},\nonumber\\
	&\quad\quad~\sum\limits_{j=1}^{T-1}(1-m\eta)^{\sum\limits_{t=j}^{T-1}\tau_{i,t}}\leq \zeta\tag{\ref{P5}{a}}\label{P5a}.
\end{align}
The optimal solutions of \eqref{P5} can be obtain by the following theorem, which are the optimized solutions of \eqref{P42}.

\textit{Theorem 4: The optimal solutions of \eqref{P5} are given by \eqref{optt} on the top of the next page.}

\begin{figure*}[!t]
	\normalsize
	\setcounter{MYtempeqncnt}{\value{equation}}
	\setcounter{equation}{34}
	\begin{equation}
		\tau_{i,t}^{*}{=}\begin{cases}
			\label{optt}
			\begin{aligned}
				&\left\lfloor \frac{\ln \big((T{-}t{+}1)^{-1}(\zeta{-}t{+}1)\big)}{\sum_{t=1}^{T}\ln \big(t^{-1}(\zeta{-}T{+}t)\big)}\tau_{\rm total}^{*}\right\rfloor
				,
				\quad\quad\quad\quad\quad 
				\tau_{\rm total}^{*}\leq \frac{\sum_{t=1}^{T}\ln \big(t^{-1}(\zeta{-}T{+}t)\big)}{\ln (T^{-1}\zeta)}\tau_{0}\\
				&1,
				\quad\quad\quad\quad\quad\quad\quad\quad\quad\quad\quad\quad\quad\quad\quad\quad\quad\quad~~
				\frac{\sum_{t=1}^{T}\ln \big(t^{-1}(\zeta{-}T{+}t)\big)}{\ln \big((T{-}t_{0}{+}1)^{-1}(\zeta{-}t_{0}{+}1)\big)}\tau_{0}\leq \tau_{\rm total}^{*} \leq \frac{\sum_{t=1}^{T}\ln \big(t^{-1}(\zeta{-}T{+}t)\big)}{\ln \big((T{-}t_{0})^{-1}(\zeta{-}t_{0})\big)}\tau_{0}\\
				&\quad\quad\quad\quad\quad\quad\quad\quad\quad\quad\quad\quad\quad\quad\quad\quad\quad\quad\quad\quad\textrm{and}~1\leq t_{0}\leq T-1, t\leq t_{0}\\
				&\left\lfloor\frac{\ln \big((T{-}t{+}1)^{-1}(\zeta{-}t{+}1)\big)}{\sum_{t=t_{0}+1}^{T}\ln \big(t^{-1}(\zeta{-}T{+}t)\big)}(\tau_{\rm total}^{*}{-}t_{0}\tau_{0})\right\rfloor,~~ \frac{\sum_{t=1}^{T}\ln \big(t^{-1}(\zeta{-}T{+}t)\big)}{\ln \big((T{-}t_{0}{+}1)^{-1}(\zeta{-}t_{0}{+}1)\big)}\tau_{0}\leq \tau_{\rm total}^{*} \leq \frac{\sum_{t=1}^{T}\ln \big(t^{-1}(\zeta{-}T{+}t)\big)}{\ln \big((T{-}t_{0})^{-1}(\zeta{-}t_{0})\big)}\tau_{0}\\
				&\quad\quad\quad\quad\quad\quad\quad\quad\quad\quad\quad\quad\quad\quad\quad\quad\quad\quad\quad\quad
				\textrm{and}~1\leq t_{0}\leq T-1, t> t_{0}\\
				&1,
				\quad\quad\quad\quad\quad\quad\quad\quad\quad\quad\quad\quad\quad\quad\quad\quad\quad\quad\quad~ \tau_{\rm total}^{*}> \frac{\sum_{t=1}^{T}\ln \big(t^{-1}(\zeta{-}T{+}t)\big)}{\ln (\zeta{-}T{+}1)}\tau_{0}
			\end{aligned}
		\end{cases}
	\end{equation}
	\hrulefill
\end{figure*}
\setcounter{equation}{\value{MYtempeqncnt}}
\setcounter{equation}{35}

\textit{Proof: See Appendix D. $\hfill\qedsymbol$}

\begin{table}[!t]
	\caption{A numerical example of problem \eqref{P42}\label{tab:table1}}
	\centering
	\begin{tabular}{c|c||c|c}
		\hline
		$(\tau_{1},\tau_{2},\tau_{3})$ & Value of \eqref{P42} & $(\tau_{1},\tau_{2},\tau_{3})$ & Value of \eqref{P42}\\
		\hline
		$(3,1,1)$ & $a^{5}+a^{2}+a$ & $(1,3,1)$ & $a^{5}+a^{4}+a$\\
		$(2,2,1)$ & $a^{5}+a^{3}+a$ & $(1,2,2)$ & $a^{5}+a^{4}+a^{2}$\\
		$(2,1,2)$ & $a^{5}+a^{3}+a^{2}$ & $\mathbf{(1,1,3)}$ &$\boldsymbol{a^{5}+a^{4}+a^{3}}$\\
		\hline
	\end{tabular}
\end{table}

Theorem 4 indicates that, the optimal rounds of local training (e.g.. $\tau_{i,t}^{*}$) only depends on its own recourse budgets, with related to $\tau_{\rm total}^{*}$ and $\tau_{0}$. Thus, $\tau_{i,t}^{*}$ can be calculated on the $i$-th device independently. Besides, from \eqref{optt}, when $\tau_{\rm total}^{*}$ and $\tau_{0}$ are fixed, the expression of $\tau_{i,t}^{*}$ is an increasing function of $t$ when $\delta_{i,E}$ is limited. It means that, at the beginning of the DFL training process, setting less rounds of local training in one iteration can achieve a better performance. This result reflects that, since there is significant divergence in the gradient at the beginning of DFL, increasing the rounds of local training in one iteration would possibly lead to further divergence, which potentially leads to the worse model performance. However, from $\sum_{t=1}^{T-1}\tau_{i,t}=\tau_{\rm total}^{*}$, we note that, smaller $\tau_{i,t}$ leads $T$ become larger, but a larger $T$ makes a smaller $\tau_{\rm total}^{*}$, which has a negative impact on improving the performance. Therefore, not all $\tau_{i,t}^{*}$ for all $t$ are 1. On the contrary, if the recourse budgets are infinity, we find that $\tau_{i,t}^{*}=1$. This conclusion is consistent with our analysis in Table I.


\subsection{Algorithm Design}

\begin{algorithm}[!t]
	\renewcommand{\algorithmicrequire}{\textbf{Input:}}
	\renewcommand{\algorithmicensure}{\textbf{Output:}}
	\caption{Adaptive DFL}
	\label{alg3}
	\begin{algorithmic}[1]
		\REQUIRE $\mathbf{w}_{0},~\delta_{i,E},~\delta_{T},~\mathcal{G}(\mathcal{V},\mathcal{E}),~\zeta$,~$\varepsilon$
		\IF{Case 1}
		\STATE For each device $i$ in parallel, calculate the energy cost of the communication with its neighbors by \eqref{eicm}.
		\STATE Applying the MST-based aggregation scheme.
		\ELSIF{Case 2 with ring topology}
		\STATE Applying the Ring-AllReduce for aggregation.
		\ELSE
		\STATE Applying the randomized gossip  for aggregation.
		\ENDIF
		\STATE Obtaining the $K_{i,t}^{*}$ and $\tau_{\rm total}^{*}$ of the applied aggregation scheme from \eqref{s1}.
		\STATE Setting $T=1$
		\WHILE{True}
		\STATE Calculate $\sum_{t=1}^{T-1}\tau_{i,t}$ and $\tau_{\rm total}$ by \eqref{optt} and \eqref{stau}.
		\IF{condition$\sum_{t=1}^{T-1}\tau_{i,t}\leq \tau_{\rm total}$}
		\STATE $T=T+1$
		\ELSE
		\STATE Break.
		\ENDIF
		\ENDWHILE
		\FOR{$t=1,2,\dots, T$}
		\STATE For each device $i$ in parallel, calculate $\tau_{i,t}$ by \eqref{optt}.
		\FOR{$j=1,2,\dots, \tau_{i,t}$}
		\STATE For each device $i$ in parallel, its local parameter vector is updated by \eqref{sgd}.
		\ENDFOR
		\STATE Aggregation parameters by the aggregation scheme.
		\ENDFOR
		\ENSURE  $\mathbf{w}_{T}$
	\end{algorithmic}  
\end{algorithm}

By combining \eqref{optt}, which is the closed-form solutions of \eqref{P42}, and Theorem 3 used to solve \eqref{P41}, we propose a DFL algorithm with a energy-saving aggregation scheme, and adaptive local training in different iterations and devices, which is summarized in Algorithm 3. Firstly, Lines 1-9 introduce the process of determining which aggregation scheme to use based on different communication conditions. Thus, each device obtains the $K_{i,t}^{*}$ and $\mathcal{E}_{t}^{*}$ by \eqref{s1} and uses them to calculate $\tau_{\rm total}^{*}$ by \eqref{stau}. Then, since $T$ is not infinity due to the limited resources, Lines 10-18 summarizes the process of obtaining $T$ of given $\delta_{i,E}$. Finally, after knowing how many iterations it needs to finish and which aggregation scheme is applied, the DFL framework begins to train. Lines 18-22 are about the whole DFL process. Specifically, by substituting $\tau_{\rm total}^{*}$ into \eqref{optt} to calculate $\tau_{i,t}^{*}$, each device can finish $\tau_{i,t}$ rounds of local training in this iteration in parallel. After all devices finish its own local training process, these parameters will be aggregated by the applied scheme which is determined before. In summary, this algorithm includes $T$ rounds of iterations, where $T$ is determined by the energy budget. Each iteration is divided in local training phase and aggregation phase. In the local training phase, since the device heterogeneity, the number of local training rounds in each device is difference. The optimized $\tau_{i,t}$ in each iteration and device obtained by \eqref{optt} enhance the model performance and utilize the recourse more efficiently. In the aggregation phase, to reduce the energy cost on communication, different graph-based aggregation schemes are applied.


\section{Simulation Results}
In this section, we evaluate the convergence, performance and energy cost of DFL with the proposed adaptive local training algorithm and energy-saving aggregation schemes by the simulations.

\subsection{System Setup}
We consider a DFL system with 20 devices. Each device trains a local machine learning model to finish some classification tasks, based on its local datasets, which is divided by the global datasets. Thus, the prediction accuracy is used to evaluate the model performance. Specifically, to make our simulation more complete, we consider three different datasets, which are MNIST, Fashion-MNIST and CIFAR-10. They correspond to grayscale image data (10 labels and 62 labels), and color image data, respectively. Based on these datasets, two different models, which are Multilayer Perceptron (MLP) and Convolutional Neural Network (CNN), are trained to finish these tasks. In addition, to guarantee that all proposed aggregation schemes can be evaluated, we consider three connection topologies among devices, which are ring, quasi-ring and grid with 2 columns, since the ring topology is the sub-graph of all these three topologies. Besides, for all devices, their transmit power $p_{t}=1$W. Their numbers of processing cycles are uniformly distributed in $[1000,3000]$ cycles/sample and their computation capacity are $2$GHz. The noise power spectral density $N_{0}$ is $ -174$dBm/Hz. In addition, some common DFL frameworks, such as Gossip Learning (GL) in \cite{gl}, balanced DFL in \cite{ra} and PENS in \cite{dfl1}, are set as three baselines, which are used to complete the comparison.

\begin{figure}[!t]
	\centering
	\subfloat[]{\includegraphics[width=2.8in]{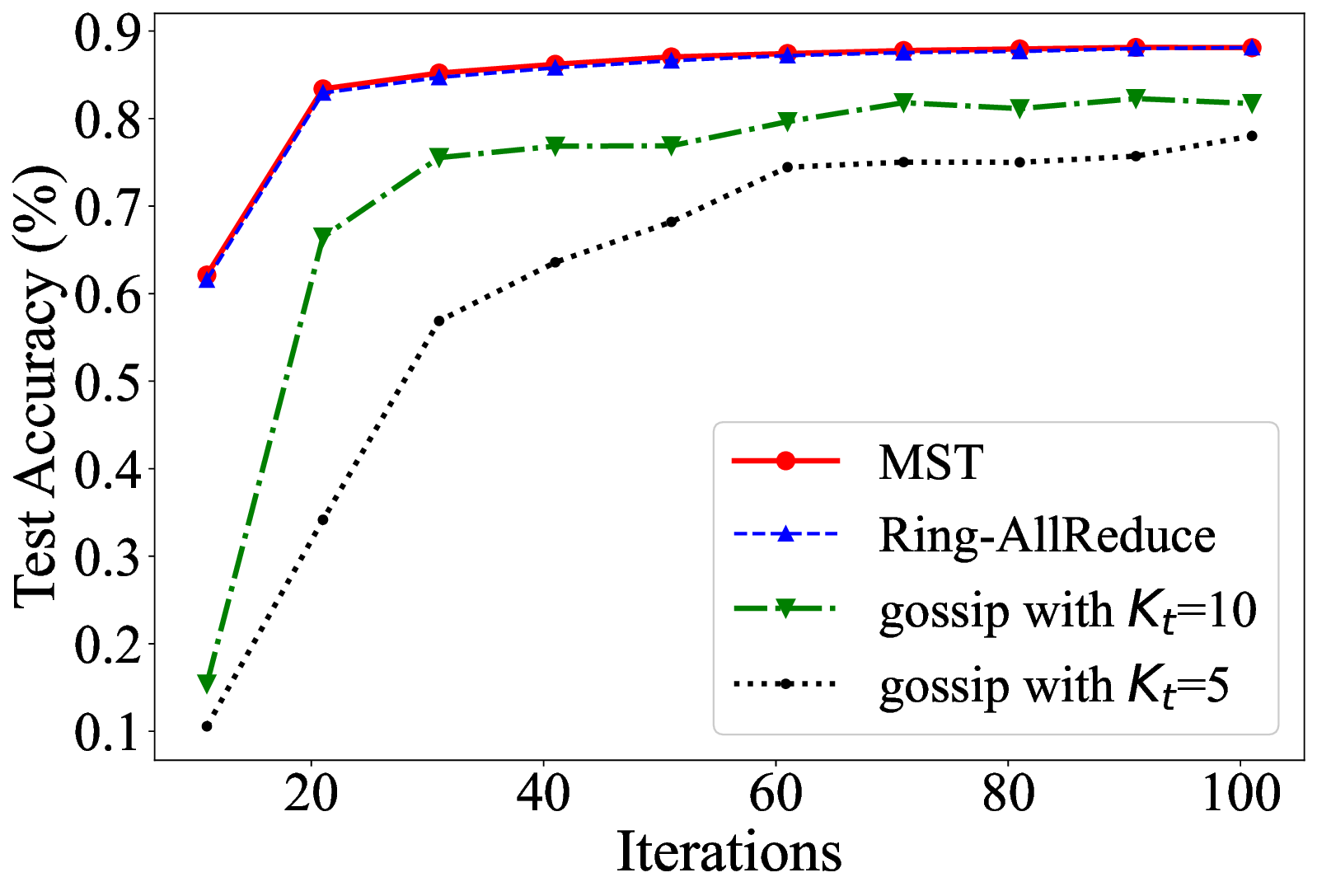}
		\label{fig2_1}}\\
	\subfloat[]{\includegraphics[width=2.8in]{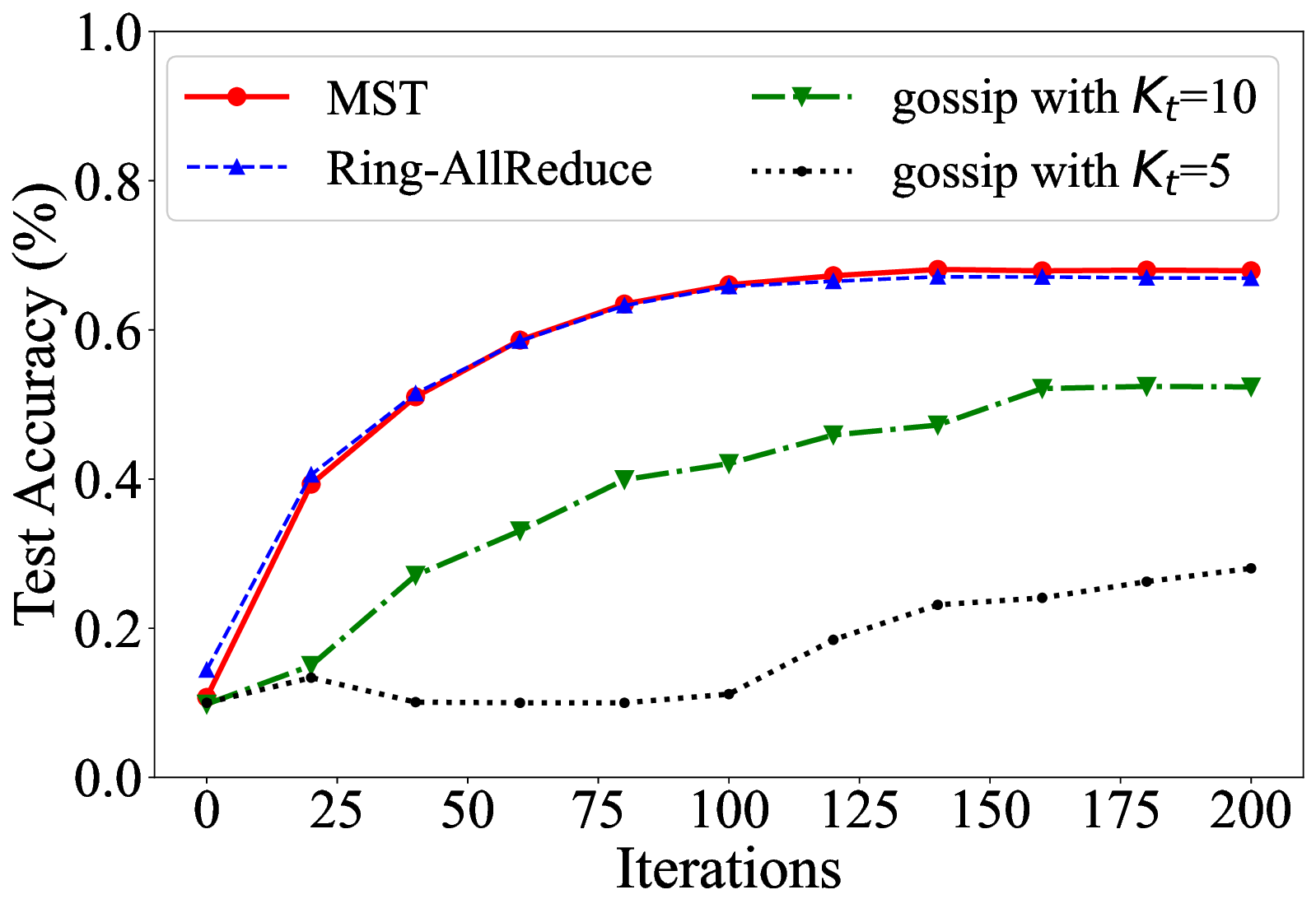}
		\label{fig2_2}}
	\caption{Simulations on Fashion-MNIST and CIFAR-10 datasets with MLP and CNN under different aggregation schemes and topolgies. (a) Test accuracy with iterations of MLP with Fashion-MNIST on the grid topology. (b) Test accuracy with iterations of CNN with CIFAR-10 on the ring topology.}
\end{figure}

\subsection{Results and Discussions}

\begin{figure}[!t]
	\centering
	\subfloat[]{\includegraphics[width=2.8in]{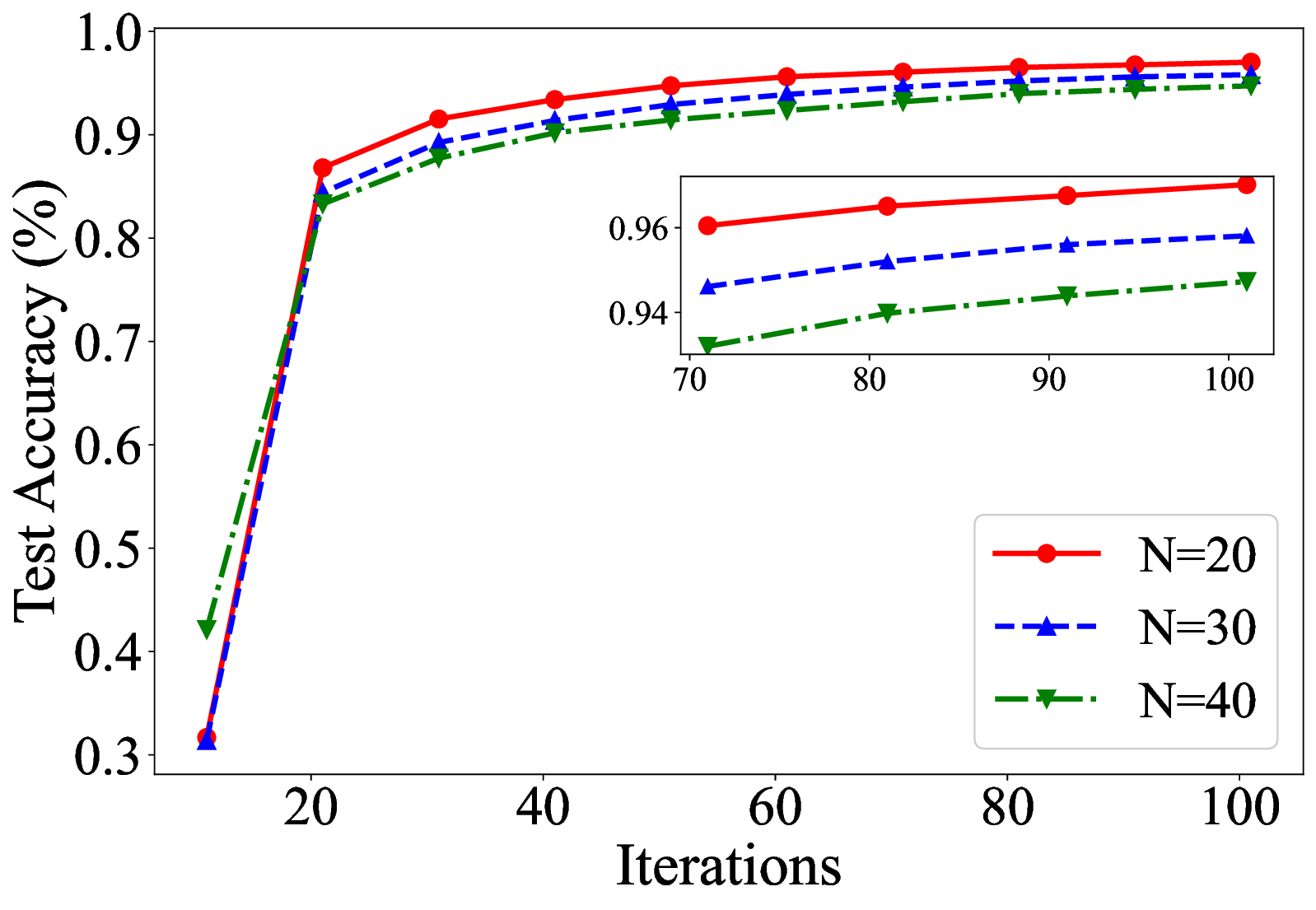}
		\label{fig2_3}}\\
	\subfloat[]{\includegraphics[width=2.8in]{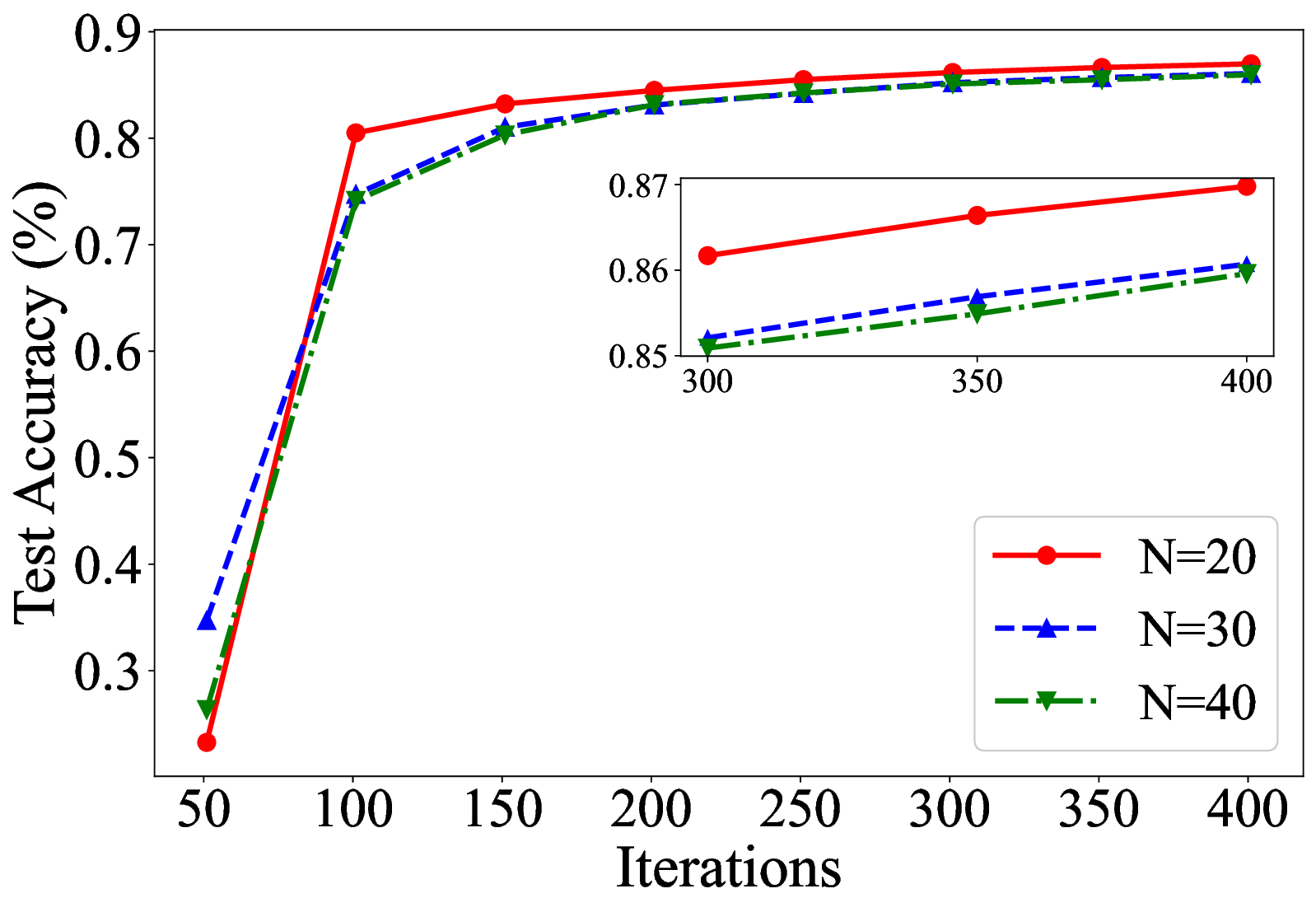}
		\label{fig2_4}}
	\caption{Simulations on MNIST and Fashion-MNIST datasets with MLP with different number of device in non-i.i.d. case. (a) Test accuracy with iterations of MLP with MNIST on different $N$. (b) Test accuracy with iterations of MLP with Fashion-MNIST on different $N$.}
\end{figure}

\textit{1) Convergence:} To verify the convergence of the proposed DFL algorithm, we train an MLP model on Fashion-MNIST and a CNN model on CIFAR-10. Both two datasets are divided into 20 parts and distributed across all devices. Furthermore, we set all these devices in the grid and completed topologies. The results are shown in Fig. 4.

Fig. 4. displays the evolution of test accuracy with increasing iterations for Fashion-MNIST with MLP on the grid topology (Fig. 2(a)) and CIFAR-10 with CNN on the ring topology (Fig. 2(b)) on various proposed aggregation schemes in the proposed DFL algorithm. Despite diverse datasets and models, consistent convergence behavior is observed. Notably, when the aggregation scheme based on the MST or the Ring-AllReduce algorithm is applied (Case 1 or Case 2 with ring topology), DFL can achieve a better performance after it converges, since there is no consensus errors in these cases (e.g., $\varepsilon=0$). Furthermore, although consensus errors can never be completely eliminated in the randomized gossip algorithm \cite{gossip}, the more communication rounds there are in one round of aggregation (e.g., a larger $K_{t}$), the smaller the consensus errors it faces. This situation can also be observed in the results. From Fig. 4., after the same rounds of DFL iterations, the randomized gossip algorithm that considers larger $K_{t}$ in the aggregation achieves higher accuracy.

Furthermore, the impact of non-i.i.d. data on the convergence can be observed in Fig. 5. The results shown in Fig. 5 are the test accuracy of MLPs on MNIST and Fashion-MNIST, which data distributions in all devices are non-i.i.d.. From these results, we note that, in the non-i.i.d. case, there are gaps in the final performance of DFL systems with different device numbers. Compared with the DFL with more devices, DFL can achieve a better performance when it has fewer devices, and this interval will not decrease as $T$ further increases after $T$ is large enough. This result confirms the analysis of Theorem 2 and is consistent with the result obtained by \eqref{gap3}.


\begin{figure}[!t]
	\centering
	\subfloat[]{\includegraphics[width=2.8in]{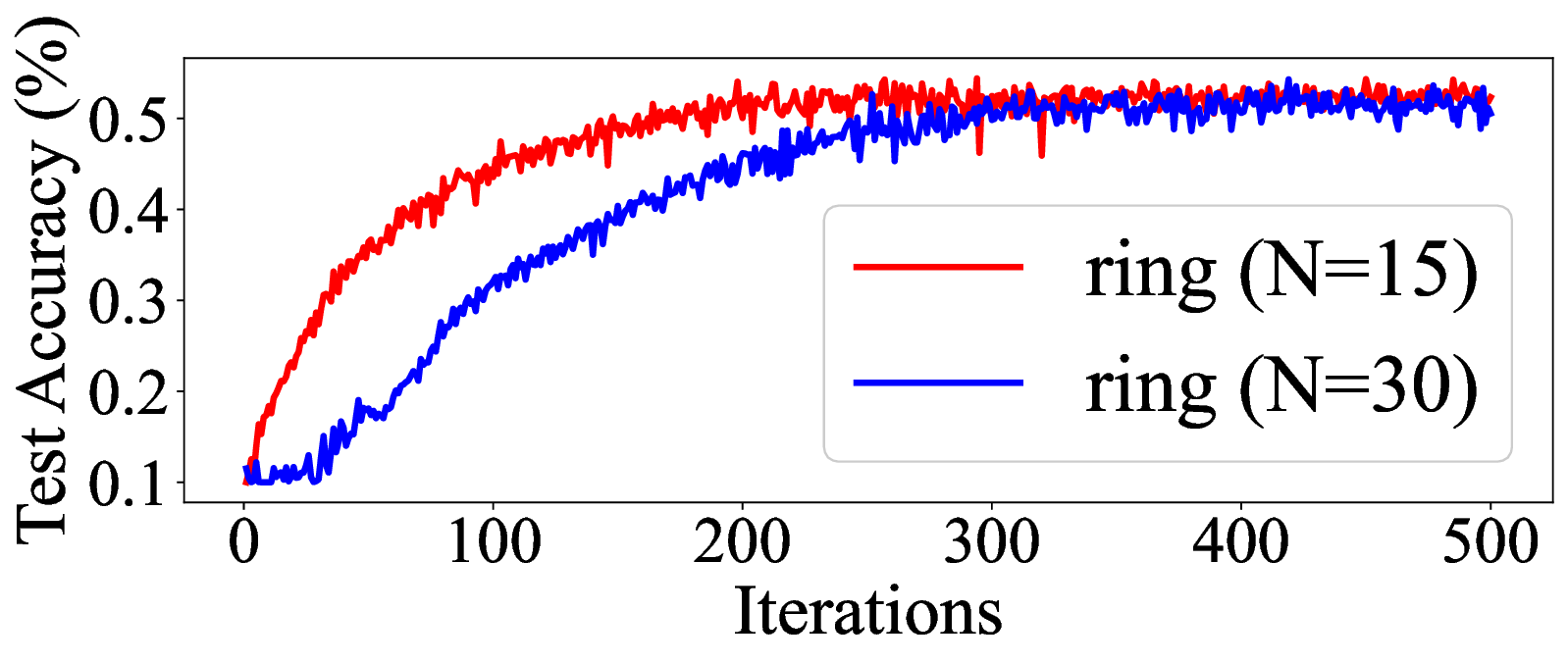}
		\label{fig3_1}}
	\\
	\subfloat[]{\includegraphics[width=2.8in]{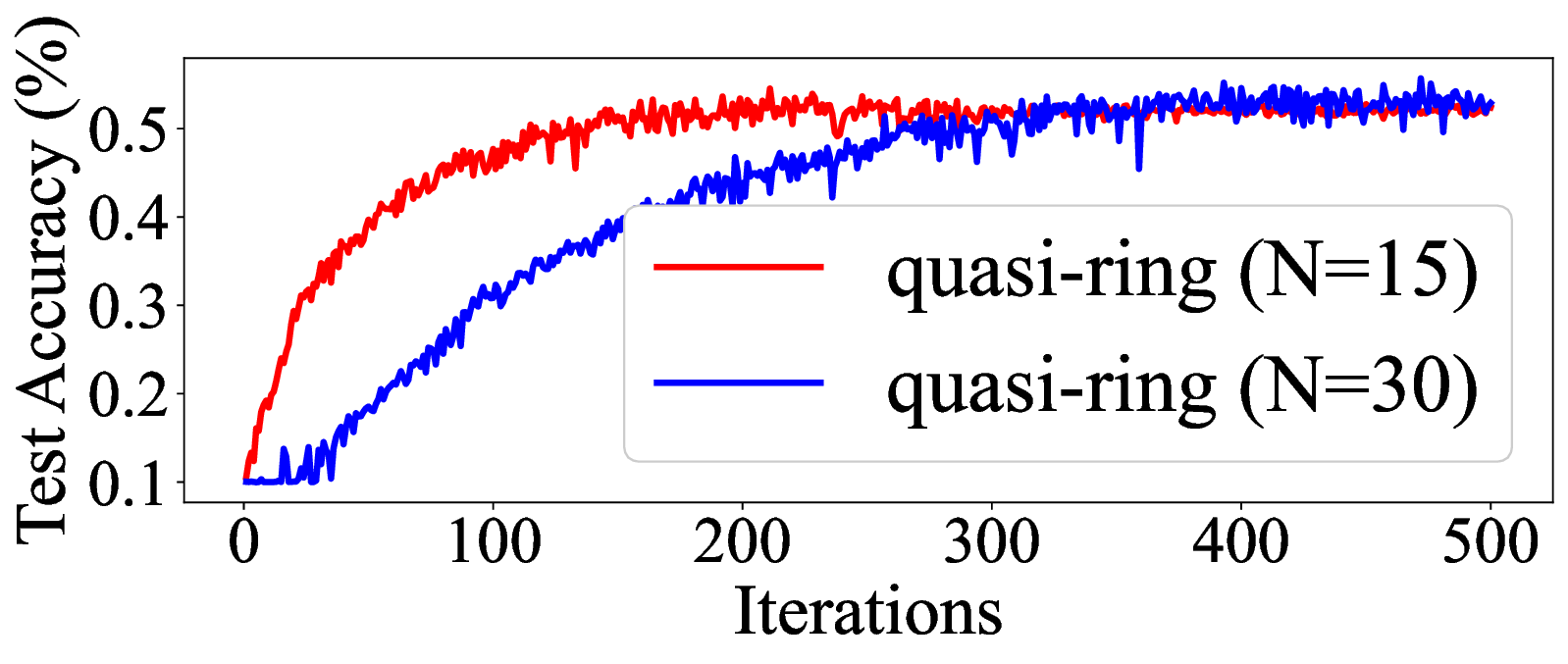}
		\label{fig3_2}}
	\caption{Simulations on CIFAR-10 datasets with CNN under different numbers of devices. (a) Test accuracy with iterations on CNN with CIFAR-10 in ring topology. (b) Test accuracy with iterations on CNN with CIFAR-10 in quasi-ring topology.} 
\end{figure}

\textit{2) Convergence Rate:} To evaluate the convergence rate of DFL with different numbers of devices under different topologies, we assume that DFL is converged after its test accuracy is more than 0.5, where the model is CNN which is trained on CIFAR-10. To reveal the impact of how number of devices and connection topology on the convergence rate, we compare the number of iterations required for the convergence of DFL with 5, 10, 15, 20 devices under the ring and quasi-ring topologies. The simulation results are shown in Fig. 6.

Examining Fig. 6(a) and 6(b), we observe the test accuracy trends with the number of iterations. Notably, fewer devices (e.g., $N=15$) lead to faster convergence compared to configurations with more devices (e.g., $N=30$), consistent with the insights from Corollary 1. 

\begin{figure}[!t]
	\centering
	\subfloat[]{\includegraphics[width=2.8in]{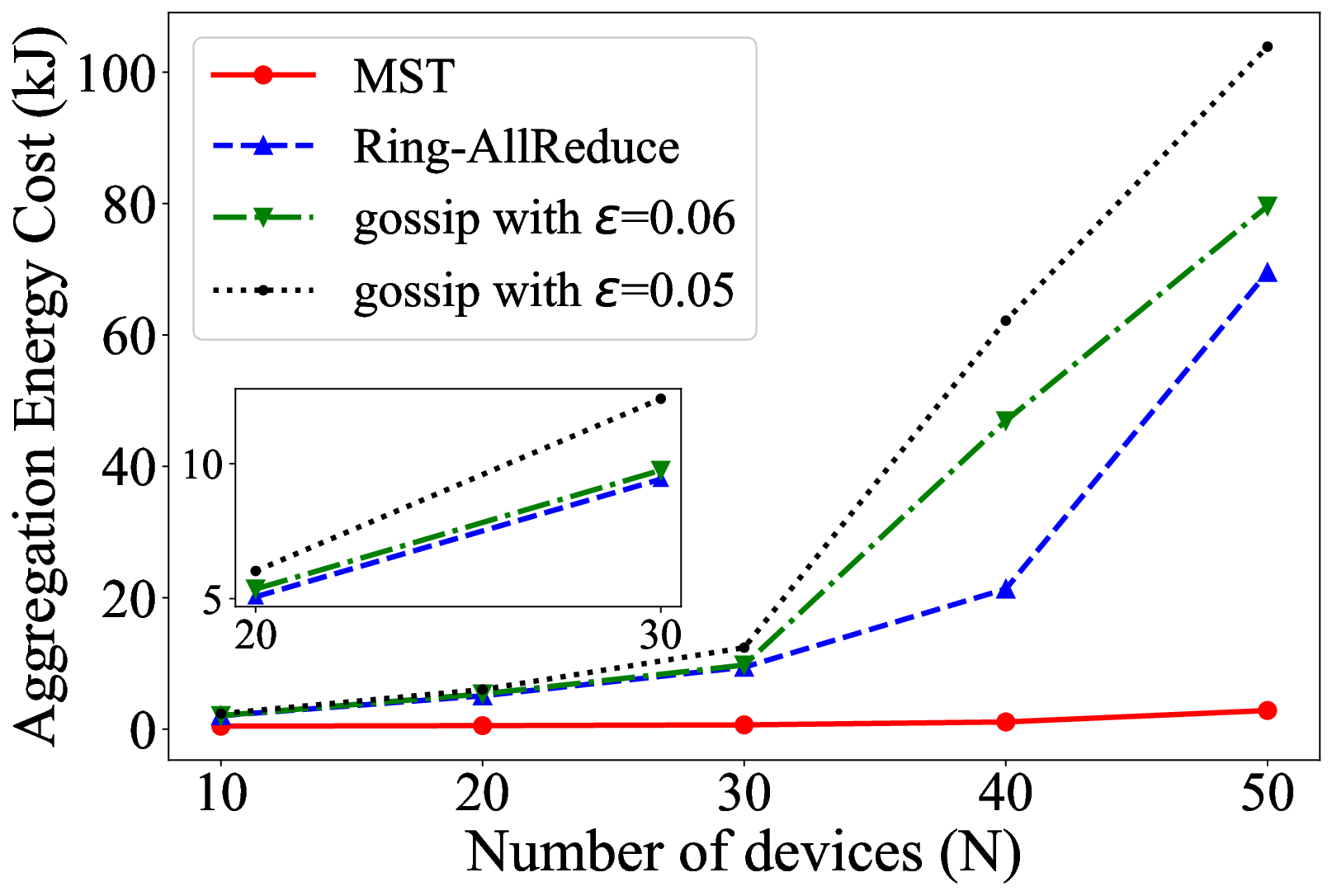}
		\label{fig6_1}}
	\\
	\subfloat[]{\includegraphics[width=2.8in]{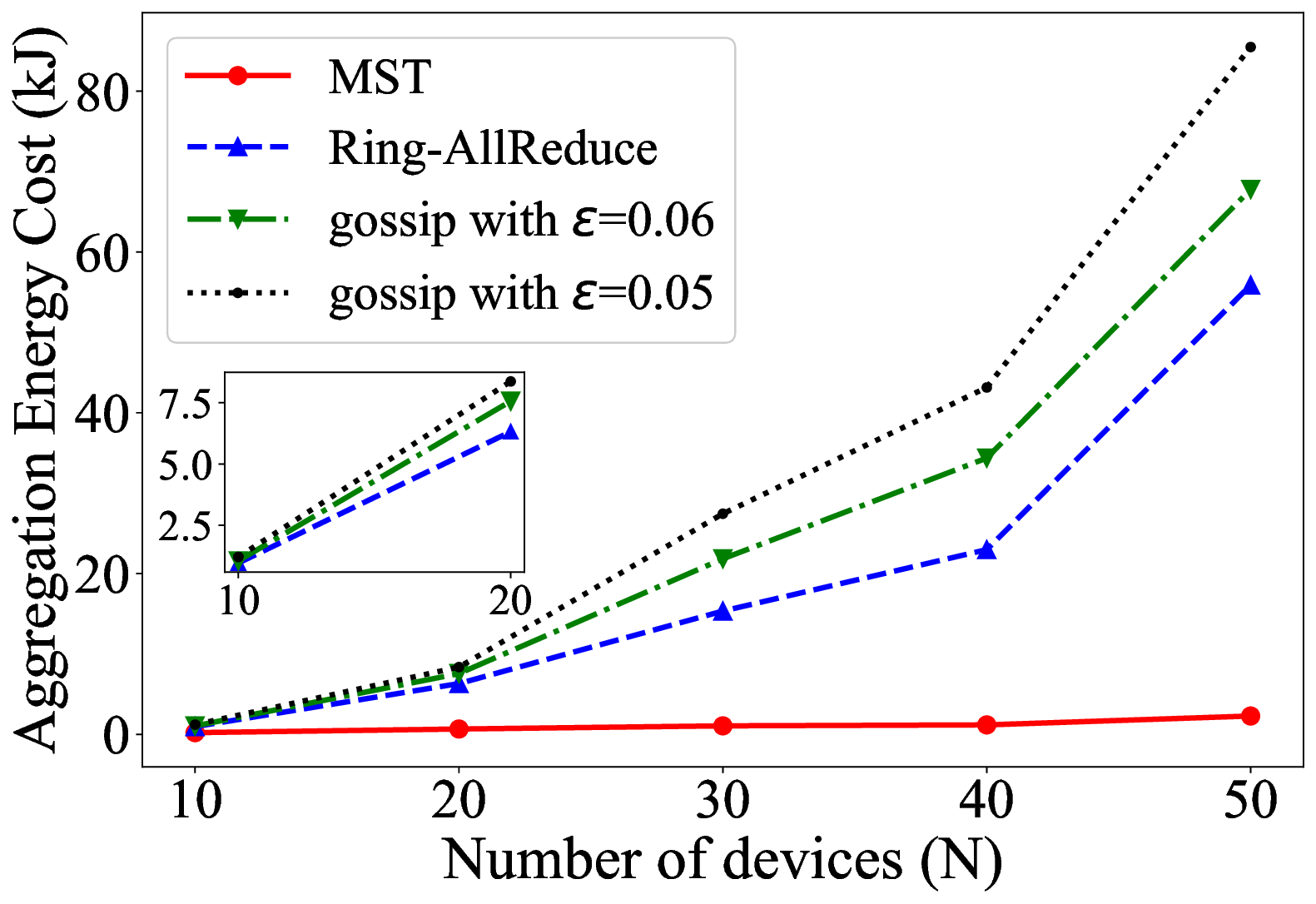}
		\label{fig6_2}}
	\caption{Simulations on Fashion-MNIST datasets with MLP under different aggregation schemes and topologies ($T=400$). (a) Total aggregation energy cost with different aggregation schemes on MLP with Fashion-MNIST in grid topology. (b) Total aggregation energy cost with different aggregation schemes on MLP with Fashion-MNIST in quasi-ring topology.}
\end{figure}

\begin{figure}[!t]
	\centering
	\subfloat[]{\includegraphics[width=2.8in]{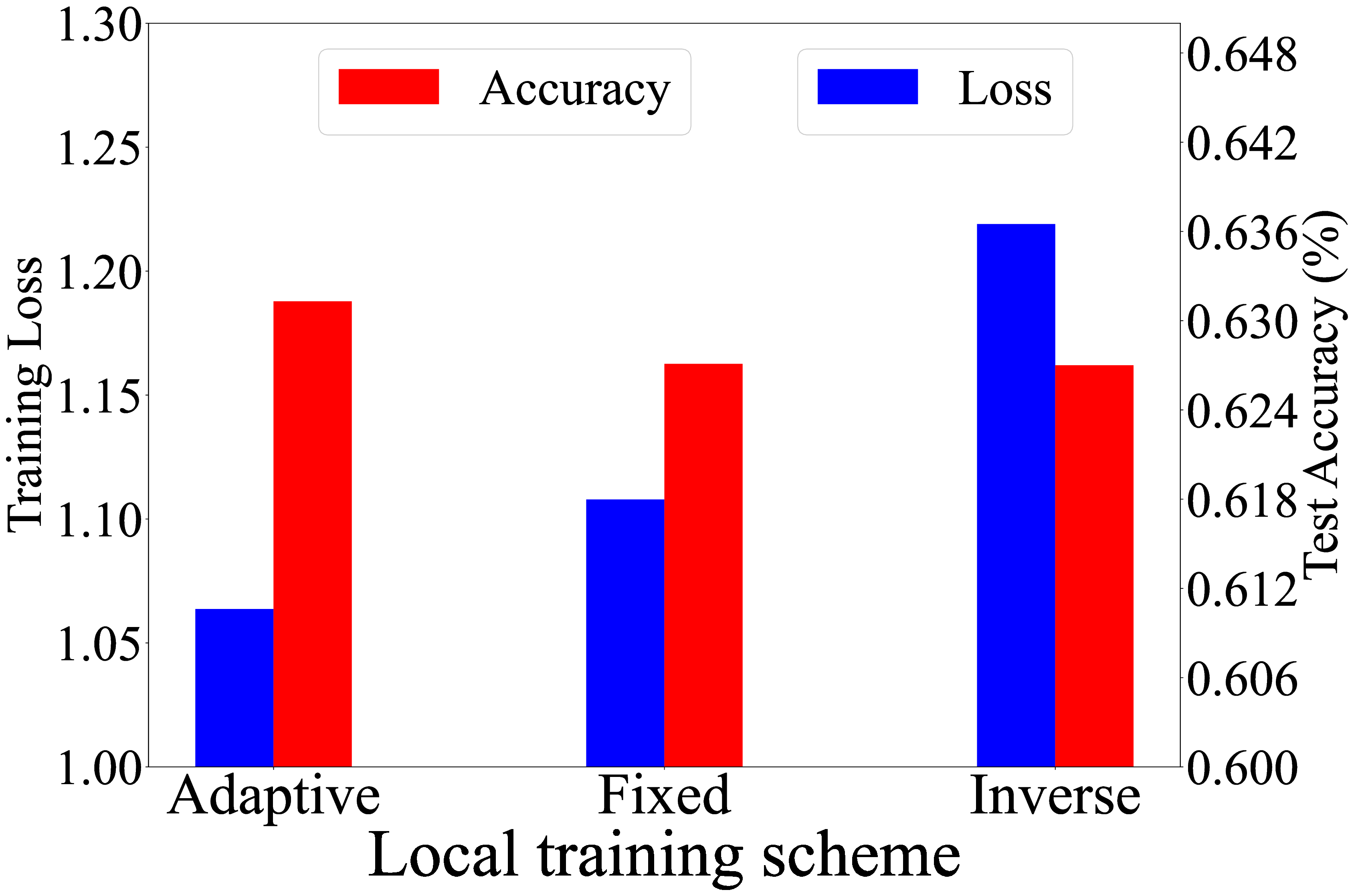}
		\label{fig5_1}}
	\\
	\subfloat[]{\includegraphics[width=2.8in]{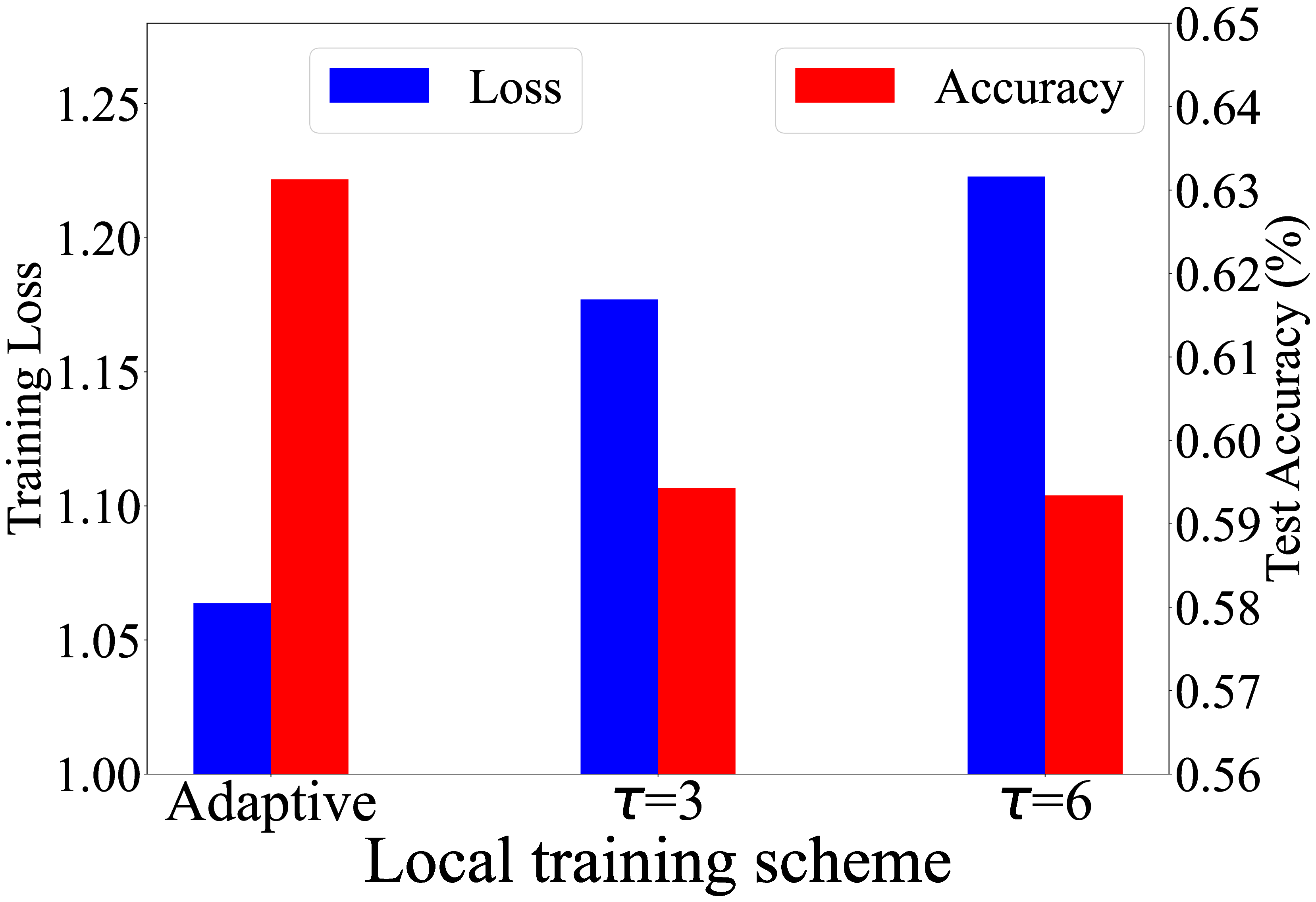}
		\label{fig5_2}}
	\caption{Simulations on CIFAR-10 datasets with CNN under number of local training rounds. (a) Training loss and test accuracy with the proposed adaptive scheme, fixed $\tau_{i,t}=4$ and inverse optimal $\tau_{i,t}$. (b) Training loss and test accuracy with the proposed adaptive scheme, fixed $\tau_{i,t}=3$ and $6$.}
\end{figure}

\begin{figure*}[!t]
	\centering
	\subfloat[]{\includegraphics[width=1.7in]{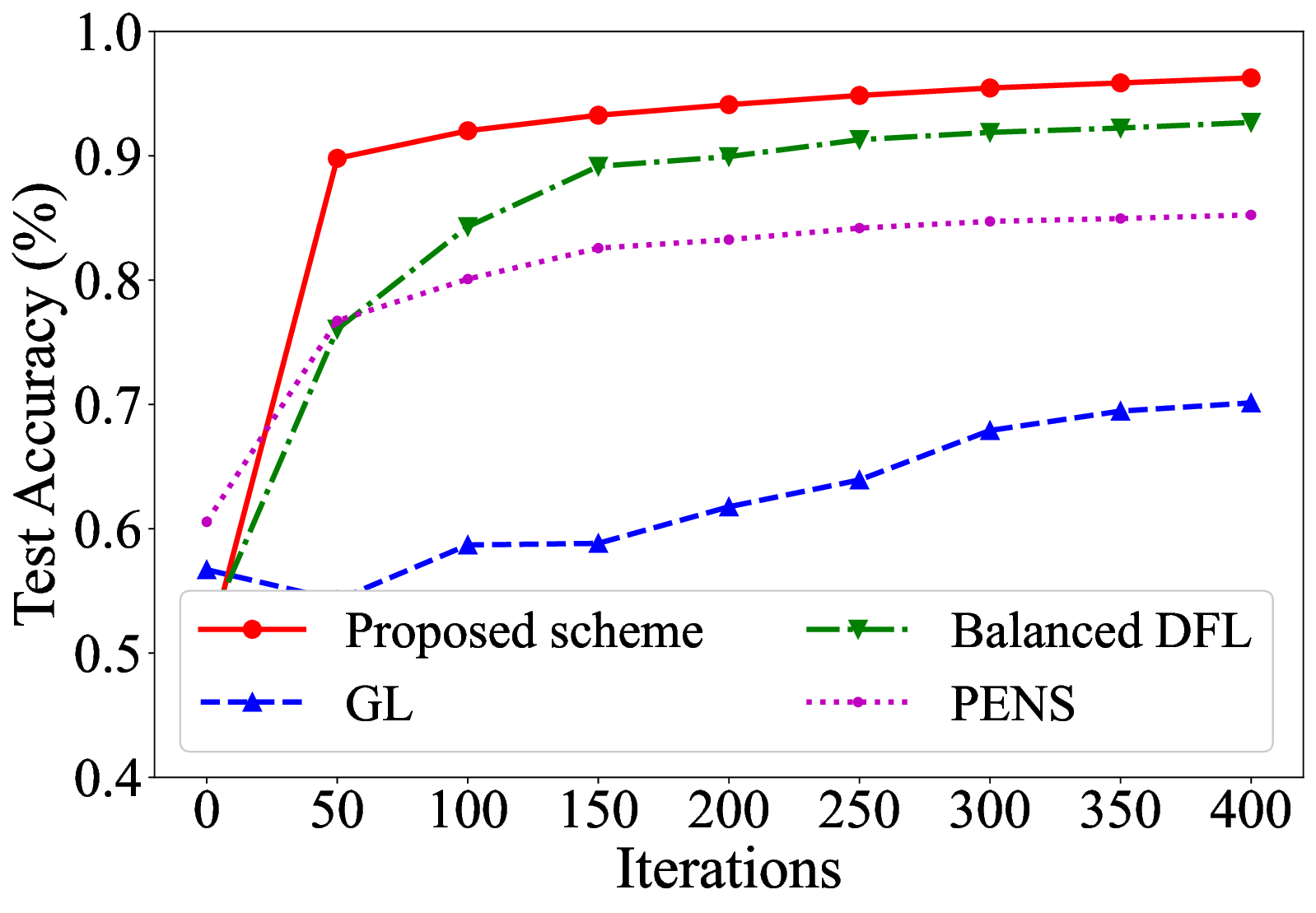}%
		\label{fig5_first_case}}
	\subfloat[]{\includegraphics[width=1.7in]{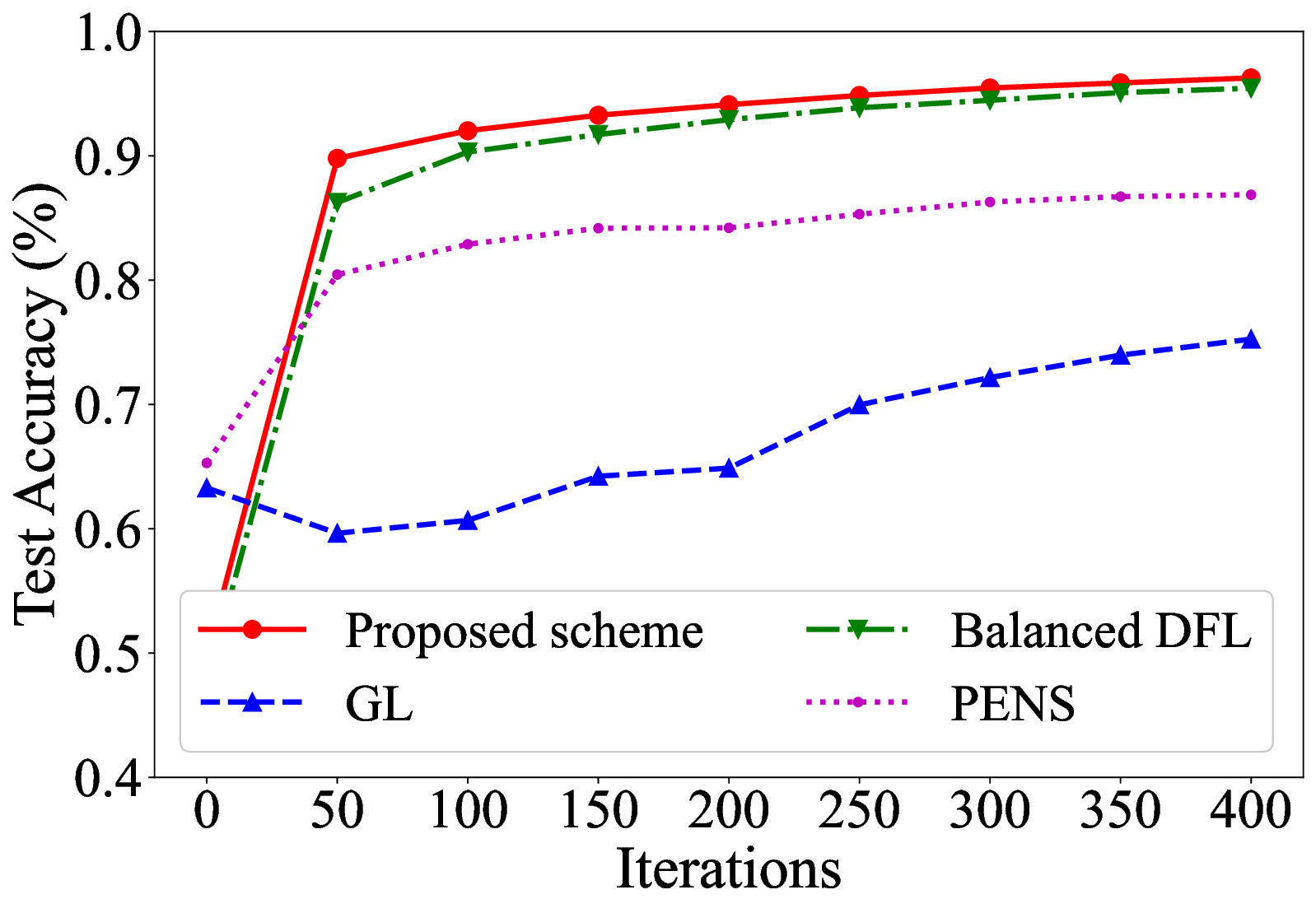}%
		\label{fig5_second_case}}
	\subfloat[]{\includegraphics[width=1.7in]{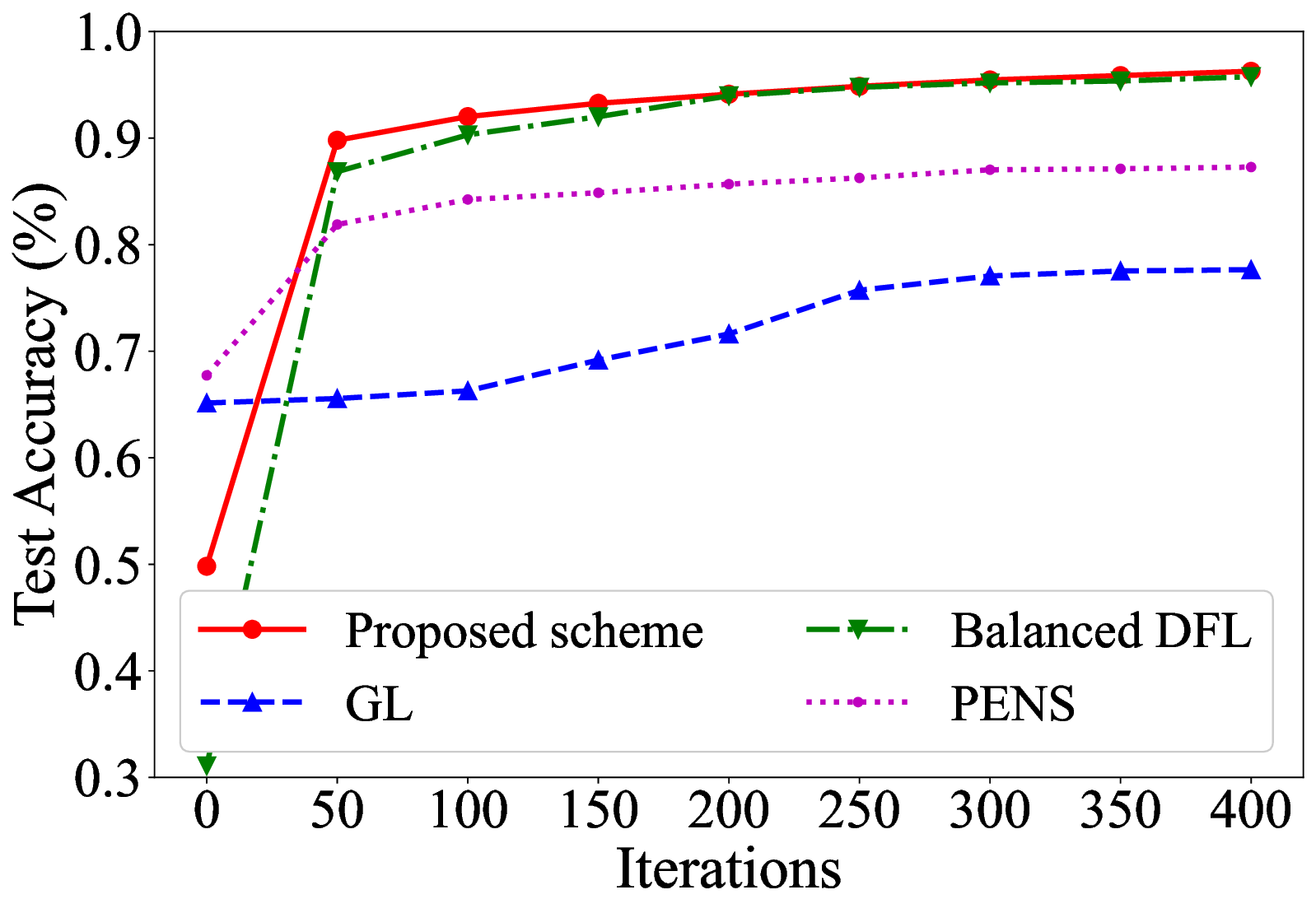}%
		\label{fig5_third_case}}
	\subfloat[]{\includegraphics[width=1.7in]{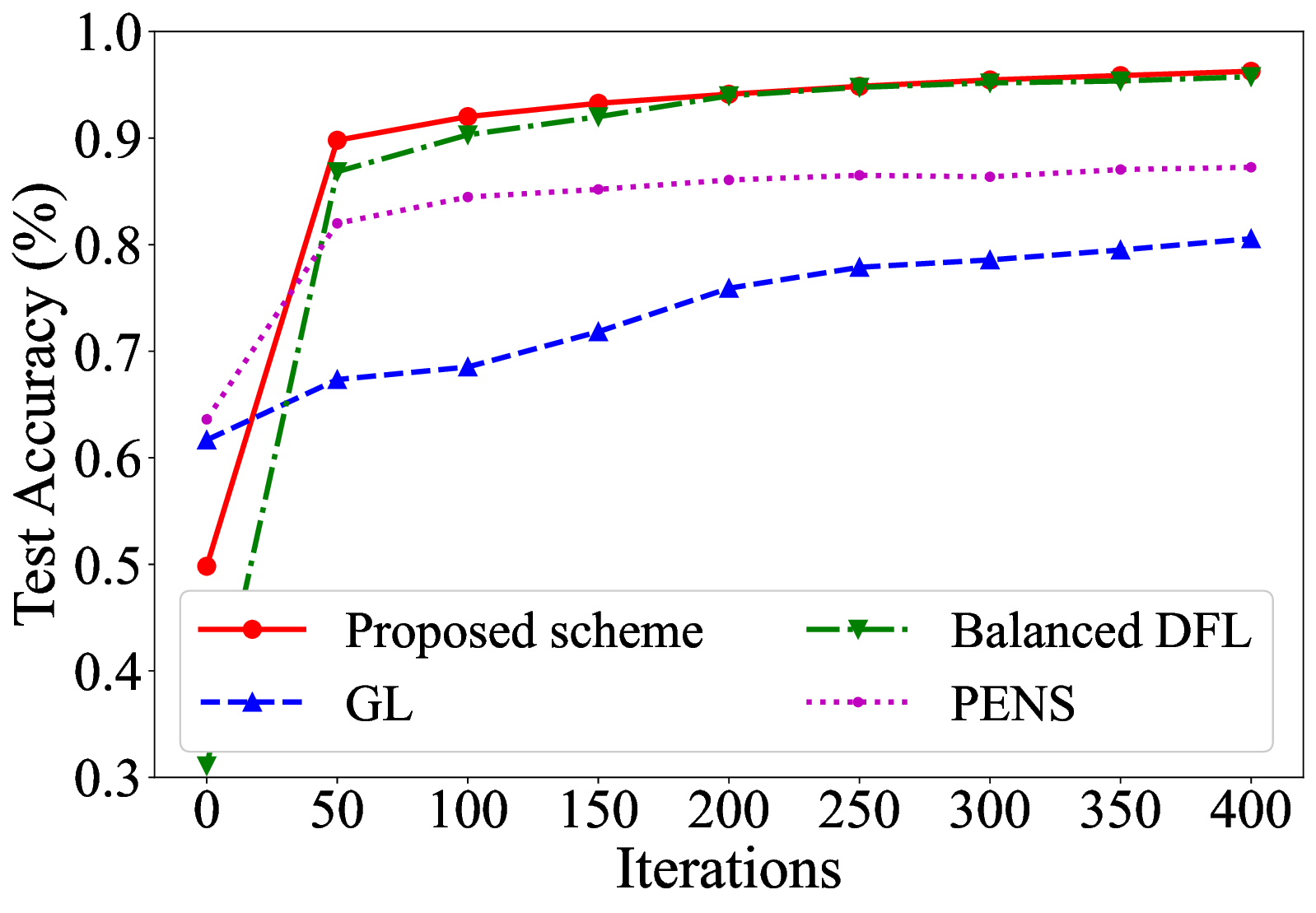}%
		\label{fig5_fourth_case}}\\\vspace{-8pt}
	\subfloat[]{\includegraphics[width=1.7in]{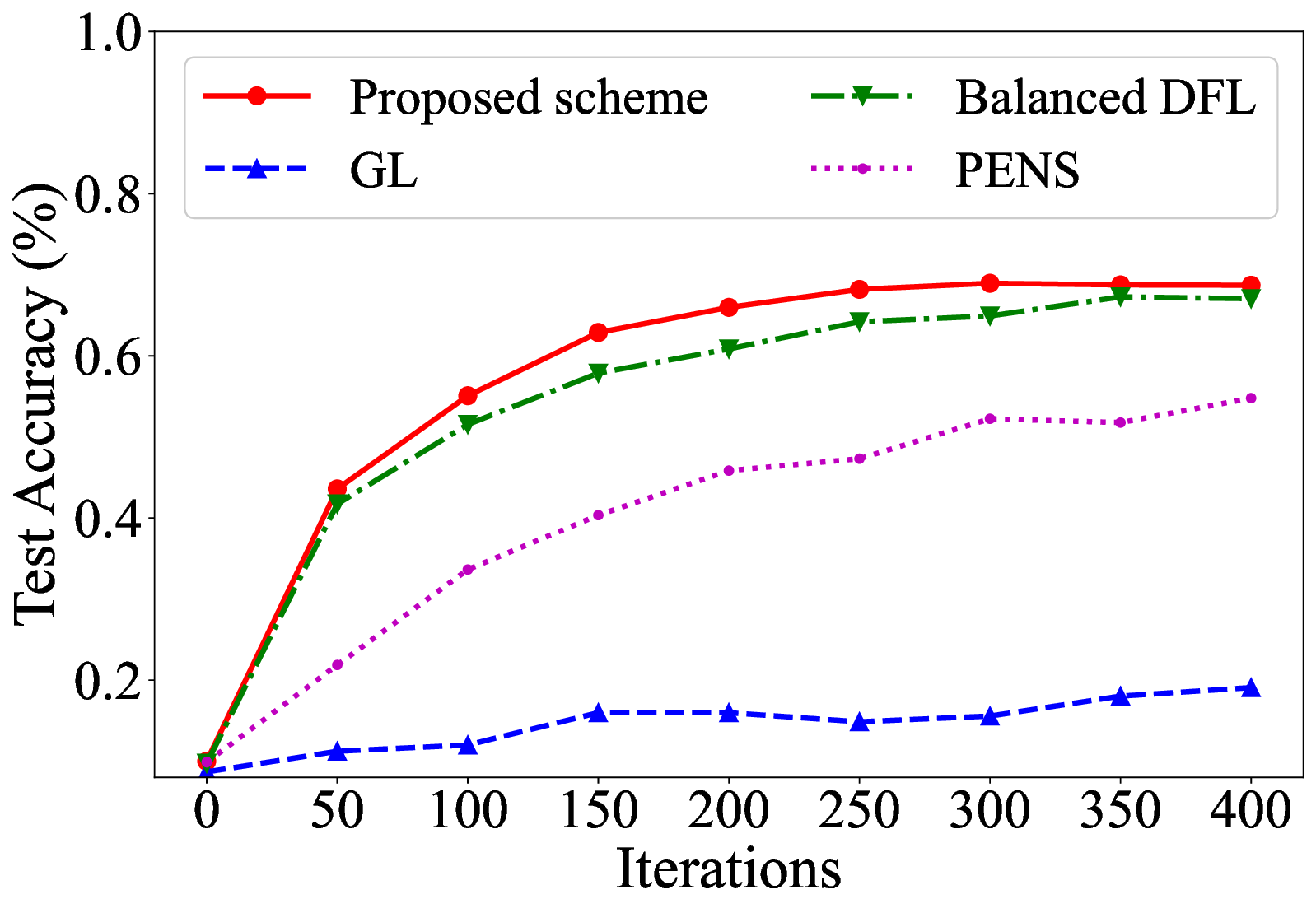}%
		\label{fig6_first_case}}
	\subfloat[]{\includegraphics[width=1.7in]{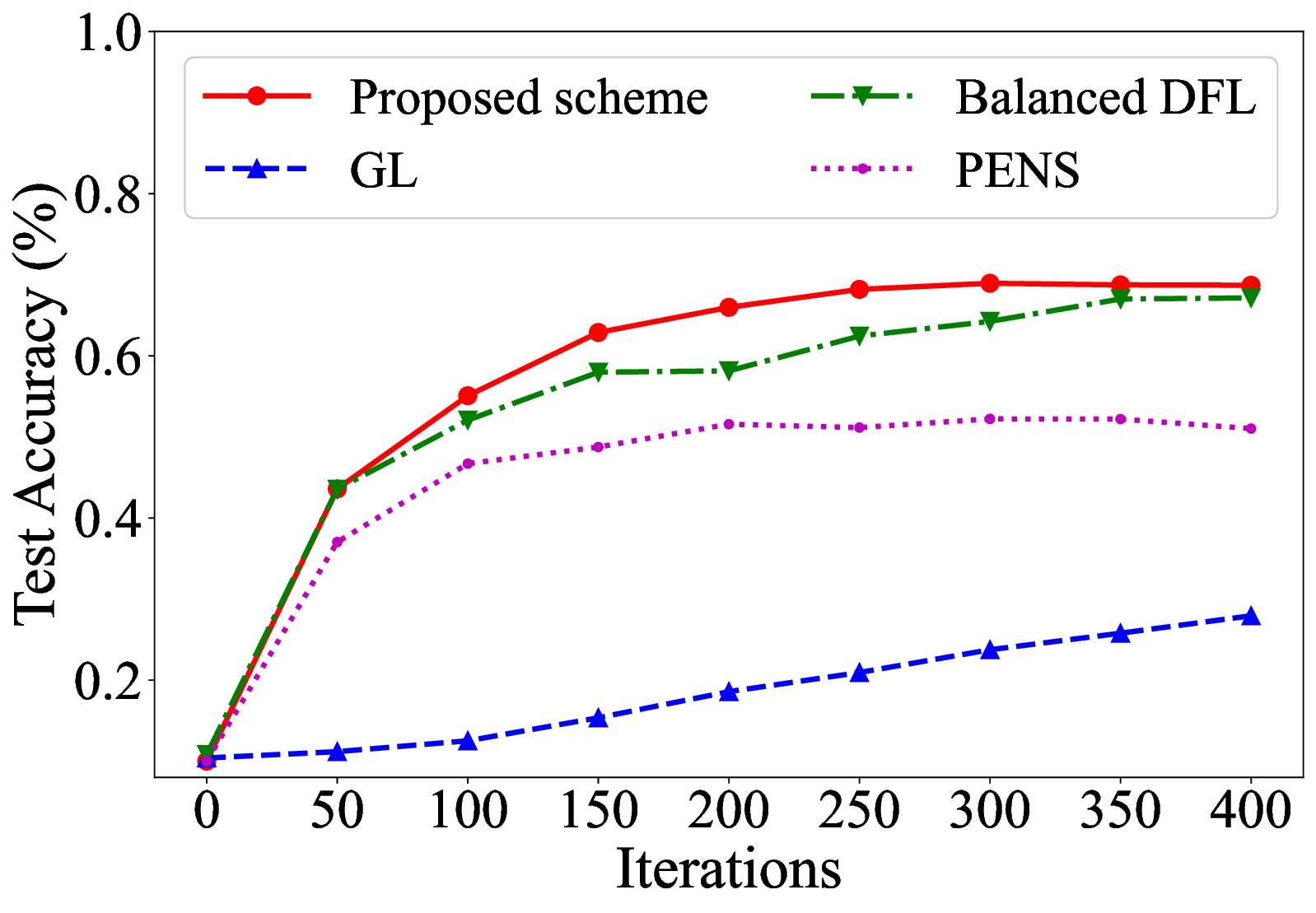}%
		\label{fig6_second_case}}
	\subfloat[]{\includegraphics[width=1.7in]{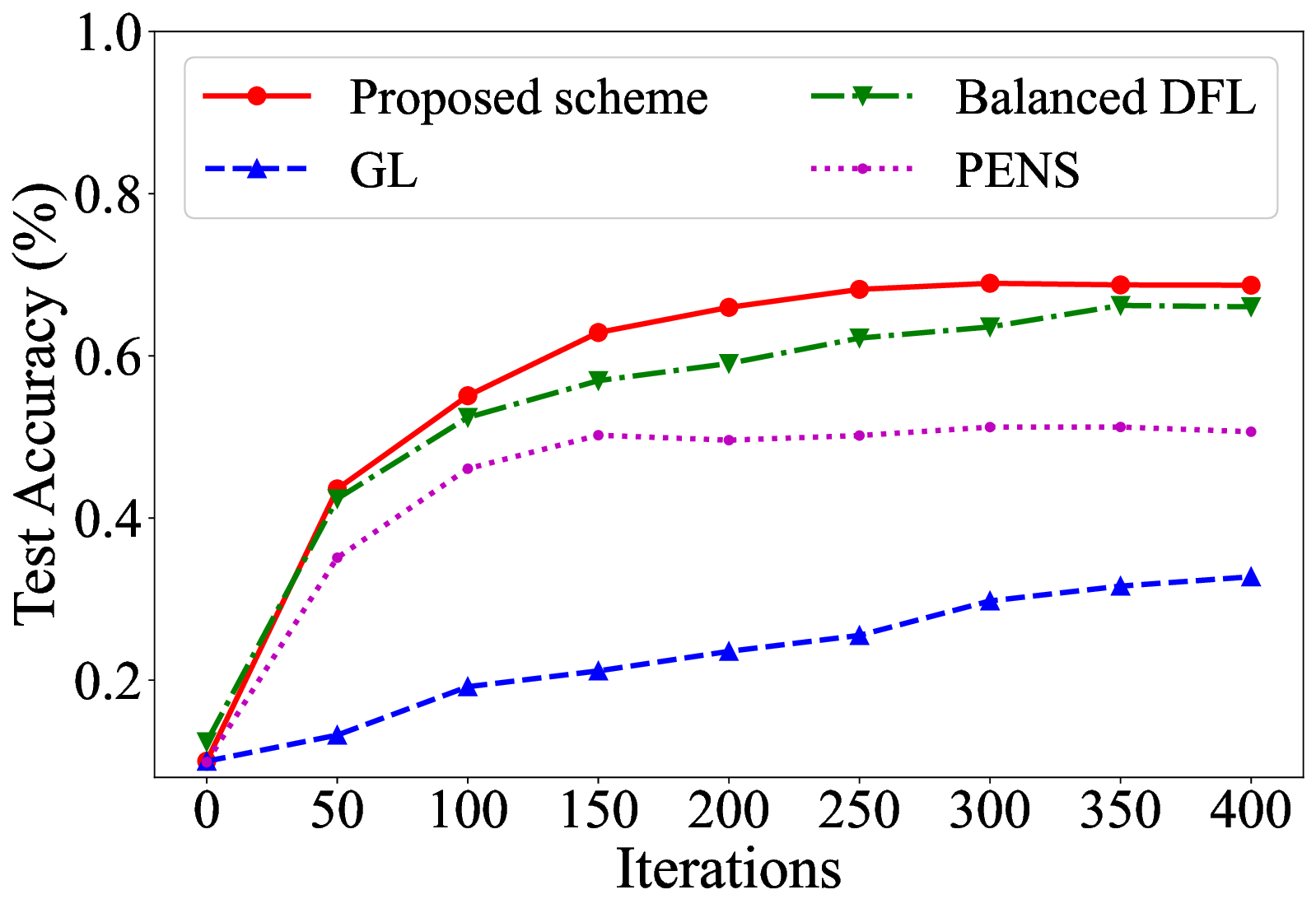}%
		\label{fi6_third_case}}
	\subfloat[]{\includegraphics[width=1.7in]{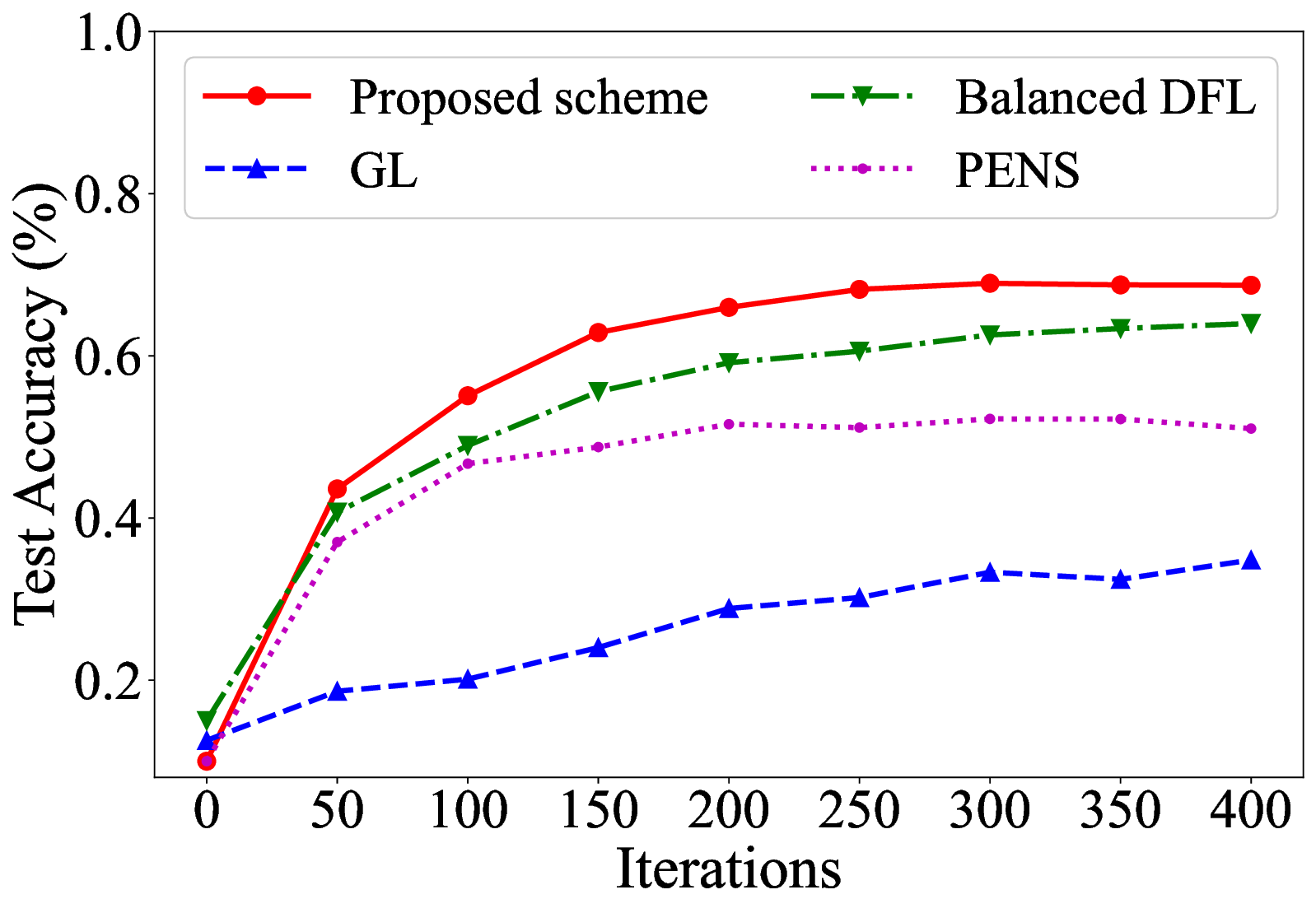}%
		\label{f6_fourth_case}}
	\caption{: Test accuracy of proposed scheme and baseline on MNIST with MLP and CIFAR-10 with CNN in different system settings. (a)-(d): MNIST with MLP in the DFL framework with 3, 5, 7 and 9 rounds of local training. (e)-(h): CIFAR-10 with CNN in the DFL framework with 3, 5, 7 and 9 rounds of local training.}
	\label{fig5_sim}\vspace{-5mm}
\end{figure*}

\textit{3) Comparison regarding the aggregation energy cost:} In this comparison, we compare the energy consumption of the proposed three aggregation algorithms. In particular, considering the communication rounds of randomized gossip algorithm is depended on the target consensus error (e.g., $\varepsilon$), we also compare two different values of $\varepsilon$ on the randomized gossip algorithm in this part. Additionally, for a comprehensive comparison, we conduct evaluations across two different topologies: grid and quasi-ring. Specifically, we train the MLP models in $400$ rounds of iteration ($T=400$) to finish the classification tasks on Fashion-MNIST datasets on the proposed DFL framework with $10,20,\cdots,50$ devices. The energy consumption of all proposed aggregation schemes are evaluated during the whole training process. The results are presented in Fig. 7.

Fig. 7. illustrates that, in the comparison to the aggregation schemes based on the randomized gossip algorithm, the proposed scheme based on the MST and Ring-AllReduce algorithms demonstrate superior energy efficiency. Results for the grid and quasi-ring topologies are presented in Fig. 7(a) and 7(b) respectively. Notably, while the energy costs of all schemes increase with the number of devices, the proposed schemes based on the MST and Ring-AllReduce consistently incur lower energy costs than the randomized gossip algorithm. In addition, to directly compare among randomized gossip algorithms with different $\varepsilon$, Fig. 7. plots the energy costs of aggregation schemes based on the randomized gossip algorithm with $\varepsilon=0.05$ and $0.06$. The result indicates that, when the randomized gossip algorithm requires a less consensus error (e.g., smaller $\varepsilon$), it needs to finish more rounds of communication in one iteration, which leads to a higher energy costs. These results are the same across the DFL with different topologies.

\textit{4) Comparison regarding the model performance:} In this part, we compare the model performance of the proposed adaptive DFL with several baselines, which can be divided into two parts as follows.

The first part is different local training schemes in the proposed DFL framework, which includes
\begin{itemize}
\item \textbf{Inverse scheme:} In this benchmark, $\tau_{i,t}$ is equal to $\tau_{i,T-t}$ calculated by \eqref{optt}. Thus, the series of $\tau_{i,t}$ in this scheme is the inverted $\tau_{i,t}$ of the proposed adaptive scheme.

\item \textbf{Fixed $\tau_{i,t}$:} In this benchmark, $\tau_{i,t}$ is fixed. We consider two cases which are $\tau_{i,t}=3$ and $\tau_{i,t}=6$. Since the recourse budgets are the same in different cases, the same $\tau_{total}=300$ in these two cases are set. Thus, the total iterations are 100 and 50 in these two cases respectively.
\end{itemize}
Another part include several proposed DFL frameworks in previous works, including
\begin{itemize}
	\item \textbf{GL \cite{gl}:} It is a DFL framework which can reduce the communication overhead between two devices by sampling a subset of the parameters.
	
	\item \textbf{Balanced DFL \cite{ra}:} It enhances the performance by balancing the rounds of local training and communication. 
	
	\item \textbf{PENS \cite{dfl1}:} It reduces the statistical heterogeneity by creating an adaptive network topology to improve the model performance. 
\end{itemize}

In these simulations, we set $\zeta=400$ in the proposed algorithm. The comparison results from the first part are presented in Fig. 8. In Fig. 8(a), three different schemes with $T=400$ are compared, and their final training loss and test accuracy are plotted. The results not only demonstrate that the proposed adaptive rounds of local training outperform others but also validate the assertions made in Theorem 4. The decrease in model performance with the inverse scheme confirms that allocating less local training rounds at the beginning of the training process is an effective approach to enhance test accuracy. Fig. 8(b) compares three schemes with the same $\tau_{\rm total}=300$. In the benchmarks with fixed $\tau_{i,t}=3$ and fixed $\tau_{i,t}=6$, their numbers of iterations are 100 and 50, respectively. This comparison highlights that allocating excessive resources to local training in each iteration, with the same resource budgets, leads to poorer performance, as devices lack sufficient communication rounds to exchange local parameters.

Fig. 9. presents comparison results between the proposed scheme and three previously introduced baselines. To provide a comprehensive comparison, we evaluatethe test accuracy using the MNIST dataset with MLP and CIFAR-10 with CNN. Unlike the proposed scheme, all baselines require the specification of the number of rounds of local training in each iteration. Therefore, we vary this number from 3, 5, 7, to 9 to observe the results under different system settings.
	
From Fig. 9, with the same number of total iterations, it is evident that while increasing the total rounds of local training can narrow the performance gap, the proposed scheme consistently outperforms the baselines. This superiority arises from its more efficient local training round allocation scheme and the ability to avoid consensus errors. Additionally, the advantages of the proposed scheme are more pronounced in complex tasks compared to simpler datasets.

\section{Conclusions}

In this paper, we have designed a DFL framework, which can utilize the limited computation and communication resources of each device to enhance the global model performance. To achieve this goal, we have formulated an optimization problem, which aims to minimize the global loss function by allocating the limited total rounds of local training into different iterations. To solve this problem, We have derived the convergence bound and convergence rate of the DFL framework with different rounds of local training among iterations and devices to relax it as a tractable problem. By solving the reformulation problem, some different aggregation schemes based on the connection graph topology of the devices are proposed to reduce the energy cost during the aggregation under the various cases with different communication conditions. In addition, the closed-form solutions for rounds of limited total rounds of local training allocation among different iterations and devices are derived, which have been provided a guideline about improving the performance of DFL by leverage the constrained resources efficiency. Our theoretical analysis is confirmed by the simulation results. The comparison results have shown that, the proposed adaptive local training scheme can achieve a better performance than some benchmarks and conventional schemes, and the graph-based aggregation scheme cost less energy than the widely-used randomized gossip algorithm.

{\appendices
	
\section{Proof of Theorem 1}
For simplicity, in the following proof, $\nabla {F}_{i,t}$ and $\nabla {F}_{i,t}^{(j)}$ are short for $\sum_{j=1}^{\tau_{i,t}}\nabla F_{i}\big(\tilde{\mathbf{w}}_{t+\frac{j}{\tau_{i,t}}},\mathcal{D}_{i}\big)$ and $\nabla F_{i}\big(\tilde{\mathbf{w}}_{t+\frac{j}{\tau_{i,t}}},\mathcal{D}_{i}\big)$, and $\mathbb{E}[\cdot]$ is short for $\mathbb{E}_{\xi_{i}\sim\mathcal{D}_{i}}[\cdot]$. To prove Theorem 1, we introduce the following lemmas.

\textit{\textbf{Lemma 1}: A lower bound of the $L_{2}$ norm of the gradient of loss function is given by}
\begin{equation}
	\begin{aligned}
		\label{L1}
		\Vert \nabla F(\mathbf{w}) \Vert_{2}^{2}\geq 2m(F(\mathbf{w})-F(\mathbf{w}')),~~\forall \mathbf{w},\mathbf{w}'
	\end{aligned}
\end{equation}

\textit{Proof: See Appendix E. $\hfill\qedsymbol$}

\textit{\textbf{Lemma 2}: An upper bound of the expected value of the $L_{2}$ norm of $\mathbf{g}_{i,t}^{(j)}$ for any $i$, $j$ and $t$ is given by}
\begin{equation}
	\begin{aligned}
		\label{L2}
		\mathbb{E}[\Vert \mathbf{g}_{i,t}^{(j)} \Vert_{2}^{2}] \leq \sigma+\Vert \nabla {F}_{i,t}^{(j)} \Vert_{2}^{2}.
	\end{aligned}
\end{equation}

\textit{Proof: See Appendix F. $\hfill\qedsymbol$}

Now, we are ready to prove Theorem 1. First, we consider a bound of $F_{i}(\tilde{\mathbf{w}}_{i,T}){-}F_{i}(\mathbf{w}_{i}^{*})$. Based on Assumption 1, we have
\begin{equation}
	\begin{aligned}
		\label{p11}
		&\mathbb{E}[F_{i}(\mathbf{w}_{i,T-1}-\eta \mathbf{g}_{i,T-1}^{(0)})]\\
		&\leq\mathbb{E}\Big[F_{i}(\mathbf{w}_{i,T-1}){-}(\nabla F_{i,T-1}^{(0)})^{T}\eta\mathbf{g}_{i,T-1}^{(0)}{+}\frac{M\eta^{2}}{2}\Vert \mathbf{g}_{i,T-1}^{(0)} \Vert_{2}^{2}\Big]\\
		&=F_{i}(\mathbf{w}_{i,T-1}){-}\eta(\nabla F_{i,T-1}^{(0)})^{T}\mathbb{E}[\mathbf{g}_{i,T-1}^{(0)}]{+}\frac{M\eta^{2}}{2}\mathbb{E}[\Vert \mathbf{g}_{i,T-1}^{(0)} \Vert_{2}^{2}]\\
		&\overset{(a)}{\leq}F_{i}(\mathbf{w}_{i,T-1}){-}\eta\Vert\nabla F_{i,T-1}^{(0)}\Vert_{2}^{2}{+}\frac{M\eta^{2}}{2}\Big( \sigma{+}\Vert \nabla {F}_{i,T-1}^{(0)} \Vert_{2}^{2}\Big)\\
		&=F_{i}(\mathbf{w}_{i,T-1}){+}\Big(\frac{M\eta^{2}}{2}{-}\eta\Big)\Vert \nabla {F}_{i,T-1}^{(0)} \Vert_{2}^{2}{+}\frac{\sigma M\eta^{2}}{2},
	\end{aligned}
\end{equation}
where step $(a)$ is based on Lemma 2. Since $\eta<\frac{1}{M}$, we have $\frac{M\eta^{2}}{2}-\eta<0$. Thus, based on Lemma 1 and $\eqref{p11}$, we have
\begin{equation}
	\begin{aligned}
		\label{p12}
		&\mathbb{E}[F_{i}(\mathbf{w}_{i,T-1}-\eta \mathbf{g}_{i,T-1}^{(0)})]\\
		&\leq F_{i}(\mathbf{w}_{i,T-1})\\
		&~~+\Big(\frac{M\eta^{2}}{2}{-}\eta\Big)2m\big( F_{i}(\mathbf{w}_{i,T-1}){-}F_{i}(\mathbf{w}_{i}^{*}) \big){+}\frac{\sigma M\eta^{2}}{2}\\
		&\overset{(b)}{\leq}F_{i}(\mathbf{w}_{i,T-1}){-}m\eta \big( F_{i}(\mathbf{w}_{i,T-1}){-}F_{i}(\mathbf{w}_{i}^{*}) \big){+}\frac{\sigma M\eta^{2}}{2}.
	\end{aligned}
\end{equation}
where step $(b)$ is based on $(\frac{M\eta^{2}}{2}{-}\eta)2m=\eta(mM\eta-2m)<\eta(m-2m)=-m\eta$ due to $\eta<\frac{1}{M}$. Thus,
\begin{equation}
	\begin{aligned}
		\label{p13}
		&\mathbb{E}[F_{i}(\mathbf{w}_{i,T-1}-\eta \mathbf{g}_{i,T-1}^{(0)})]-F_{i}(\mathbf{w}_{i}^{*})\\
		&\leq F_{i}(\mathbf{w}_{i,T-1}){-}F_{i}(\mathbf{w}_{i}^{*}){-}m\eta \big( F_{i}(\mathbf{w}_{i,T-1}){-}F_{i}(\mathbf{w}_{i}^{*}) \big){+}\frac{\sigma M\eta^{2}}{2}\\
		&=(1-m\eta)\big( F_{i}(\mathbf{w}_{i,T-1}){-}F_{i}(\mathbf{w}_{i}^{*}) \big){+}\frac{\sigma M\eta^{2}}{2}.
	\end{aligned}
\end{equation}
Then, repeating the manipulation in \eqref{p13}, we can get
\begin{equation}
	\begin{aligned}
		\label{p14}
		\mathbb{E}&[F_{i}(\tilde{\mathbf{w}}_{i,T})-F_{i}(\mathbf{w}_{i}^{*})]\\
		&\leq (1-m\eta)^{\tau_{i,T-1}}\big( F_{i}(\mathbf{w}_{i,T-1})-F_{i}(\mathbf{w}_{i}^{*}) \big)\\
		&~~+\Big(1+(1-m\eta)+\cdots+(1-m\eta)^{\tau_{i,T-1}}\Big)\frac{\sigma M\eta^{2}}{2}\\
		&=(1-m\eta)^{\tau_{i,T-1}}\big( F_{i}(\mathbf{w}_{i,T-1})-F_{i}(\mathbf{w}_{i}^{*}) \big)\\
		&~~+\frac{1-(1-m\eta)^{\tau_{i,T-1}-1}}{m\eta}\cdot\frac{\sigma M\eta^{2}}{2}\\
		&\leq (1-m\eta)^{\tau_{i,T-1}}\big( F_{i}(\mathbf{w}_{i,T-1})-F_{i}(\mathbf{w}_{i}^{*}) \big)+\frac{\sigma M\eta}{2m}.
	\end{aligned}
\end{equation}
Thus, the bound of $F_{i}(\mathbf{w}_{i,T})-F_{i}(\mathbf{w}_{i}^{*})$ can be written as
\begin{equation}
	\begin{aligned}
		\label{p15}
		&F_{i}(\mathbf{w}_{i,T})-F_{i}(\mathbf{w}_{i}^{*})\\
		&=\mathbb{E}[F_{i}(\mathbf{w}_{i,T})-F_{i}(\tilde{\mathbf{w}}_{i,T})+F_{i}(\tilde{\mathbf{w}}_{i,T})-F_{i}(\mathbf{w}_{i}^{*})]\\
		&\leq\mathbb{E}[F_{i}(\mathbf{w}_{i,T})-F_{i}(\tilde{\mathbf{w}}_{i,T})]\\
		&~~+(1{-}m\eta)^{\tau_{i,T-1}}\big( F_{i}(\mathbf{w}_{i,T-1}){-}F_{i}(\mathbf{w}_{i}^{*}) \big)+\frac{\sigma M\eta}{2m}\\
		&\leq \mathbb{E}[F_{i}(\mathbf{w}_{i,T})-F_{i}(\tilde{\mathbf{w}}_{i,T})]+\frac{\sigma M\eta}{2m}\\
		&~~+(1{-}m\eta)^{\tau_{i,T-1}}\Big(\mathbb{E}[F_{i}(\mathbf{w}_{i,T-1})-F_{i}(\tilde{\mathbf{w}}_{i,T-1})]\\
		&~~~~+(1{-}m\eta)^{\tau_{i,T-2}}\big( F_{i}(\mathbf{w}_{i,T-2}){-}F_{i}(\mathbf{w}_{i}^{*}) \big)+\frac{\sigma M\eta}{2m}\Big)\\
		&\leq (1{-}m\eta)^{\sum\limits_{t=1}^{T-1}\tau_{i,t}}\big(F_{i}(\mathbf{w}_{i,1}){-}F_{i}(\mathbf{w}_{i}^{*})\big)\\
		&~~+\frac{\sigma M\eta}{2m}\Big(1{+}(1{-}m\eta)^{\tau_{i,T-1}}{+}\cdots{+}(1{-}m\eta)^{\sum\limits_{t=1}^{T-1}\tau_{i,t}}\Big)\\
		&~~+\Big(\mathbb{E}[F_{i}(\mathbf{w}_{i,T}){-}F_{i}(\tilde{\mathbf{w}}_{i,T})]\\
		&~~~~~~+(1{-}m\eta)^{\tau_{i,T-1}}\mathbb{E}[F_{i}(\mathbf{w}_{i,T-1}){-}F_{i}(\tilde{\mathbf{w}}_{i,T-1})]\\
		&~~~~~~+\cdots+(1{-}m\eta)^{\sum\limits_{t=1}^{T-1}\tau_{i,t}}\mathbb{E}[F_{i}(\mathbf{w}_{i,1}){-}F_{i}(\tilde{\mathbf{w}}_{i,1})]\Big)\\
		&=(1{-}m\eta)^{\sum\limits_{t=1}^{T-1}\tau_{i,t}}\big(F_{i}(\mathbf{w}_{i,1}){-}F_{i}(\mathbf{w}_{i}^{*})\big){+}\frac{\sigma M\eta}{2m}\\
		&~~\Big(1{+}\sum\limits_{j=1}^{T-1}(1{-}m\eta)^{\sum\limits_{t=j}^{T-1}\tau_{i,j}}\Big){+}\Big(\mathbb{E}[F_{i}(\mathbf{w}_{i,T}){-}F_{i}(\tilde{\mathbf{w}}_{i,T})]\\
		&~~+\sum\limits_{j=1}^{T-1}(1{-}m\eta)^{\sum\limits_{t=j}^{T-1}\tau_{i,j}}\mathbb{E}[F_{i}(\mathbf{w}_{i,j}){-}F_{i}(\tilde{\mathbf{w}}_{i,j})]\Big)\\
		&\leq (1{-}m\eta)^{\sum\limits_{t=1}^{T-1}\tau_{i,t}}\big(F_{i}(\mathbf{w}_{i,1}){-}F_{i}(\mathbf{w}_{i}^{*})\big){+}\frac{\sigma M\eta}{2m}\\
		&~\big(1{+}\sum\limits_{j=1}^{T-1}(1{-}m\eta)^{\sum\limits_{t=j}^{T-1}\tau_{i,j}}\big){+}\sum\limits_{t=1}^{T}\mathbb{E}[F_{i}(\mathbf{w}_{i,t}){-}F_{i}(\tilde{\mathbf{w}}_{i,t})].
	\end{aligned}
\end{equation}
Thus, based on the definition of $F(\mathbf{w}_{T})$ and $F(\mathbf{w}^{*})$ shown in \eqref{gf}, we have
\begin{equation}
	\begin{aligned}
		\label{p16}
		F&(\mathbf{w}_{T}){-}F(\mathbf{w}^{*})\\
		&\leq\frac{1}{N}\sum_{i=1}^{N}(1{-}m\eta)^{\sum\limits_{t=1}^{T-1}\tau_{i,t}}\big(F_{i}(\mathbf{w}_{i,1}){-}F_{i}(\mathbf{w}_{i}^{*})\big)\\
		&~~+\frac{1}{N}\sum_{i=1}^{N}\frac{\sigma M\eta}{2m}\Big(1{+}\sum\limits_{j=1}^{T-1}(1{-}m\eta)^{\sum\limits_{t=j}^{T-1}\tau_{i,j}}\Big)\\
		&~~+\sum\limits_{t=1}^{T}\mathbb{E}[F(\mathbf{w}_{t}){-}F(\tilde{\mathbf{w}}_{t})].
	\end{aligned}
\end{equation}
This completes the proof.

\section{Proof of Theorem 2}

According to the Assumption 1, we can obtain
\begin{equation}
	\begin{aligned}
		\label{p17}
		\mathbb{E}&[F(\mathbf{w}_{T}){-}F(\tilde{\mathbf{w}}_{T})]\\
		&\leq \mathbb{E}\Big[\nabla F(\tilde{\mathbf{w}}_{T})^{T}(\mathbf{w}_{T}{-}\tilde{\mathbf{w}}_{T}){+}\frac{M}{2}\Vert \mathbf{w}_{T}{-}\tilde{\mathbf{w}}_{T} \Vert_{2}^{2}\Big]\\
		&\overset{(c)}{\leq}\mathbb{E}\Big[\frac{\gamma}{2}\Vert\nabla F(\tilde{\mathbf{w}}_{T})\Vert_{2}^{2}+\Big(\frac{1}{2\gamma}+\frac{M}{2}\Big)\Vert \mathbf{w}_{T}{-}\tilde{\mathbf{w}}_{T} \Vert_{2}^{2}\Big]\\
		&=\frac{\gamma}{2}\mathbb{E}[\Vert\nabla F(\tilde{\mathbf{w}}_{T})\Vert_{2}^{2}]+\Big(\frac{1}{2\gamma}+\frac{M}{2}\Big)\mathbb{E}[\Vert \mathbf{w}_{T}{-}\tilde{\mathbf{w}}_{T} \Vert_{2}^{2}],
	\end{aligned}
\end{equation}
where step $(c)$ follows from Young's inequality $\pm 2a^{T}b\leq\gamma\Vert a \Vert^{2}+\gamma^{-1}\Vert b \Vert^{2}$ for any $\gamma>0$. On one hand, based on Assumption 2 to 4, $\mathbb{E}[\Vert\nabla F(\tilde{\mathbf{w}}_{T})\Vert_{2}^{2}]$ is bounded by
\begin{equation}
	\begin{aligned}
			\label{p18}
			\mathbb{E}[\Vert\nabla& F(\tilde{\mathbf{w}}_{T})\Vert_{2}^{2}]=\mathbb{E}\Big[\Big\Vert\frac{1}{N}\sum\limits_{i=1}^{N}\nabla F_{i}(\tilde{\mathbf{w}}_{i,T})\Big\Vert_{2}^{2}\Big]\\
			&=\mathbb{E}\Big[\Big\Vert\frac{1}{N}\sum\limits_{i=1}^{N}\nabla F_{i,T}-\mathbf{g}_{i,T}+\mathbf{g}_{i,T}\Big\Vert_{2}^{2}\Big]\\
			&\leq \mathbb{E}\Big[\Big(\Big\Vert\frac{1}{N}\sum\limits_{i=1}^{N}\nabla F_{i,T}-\mathbf{g}_{i,T}\Big\Vert_{2}^{2}+\Vert\mathbf{g}_{i,T}\Vert_{2}^{2}\Big)^{2}\Big]\\
			&\overset{(d)}{\leq}\mathbb{E}\Big[\Big\Vert\frac{1}{N}\sum\limits_{i=1}^{N}\nabla F_{i,T}-\mathbf{g}_{i,T}\Big\Vert_{2}^{2}+\Vert\mathbf{g}_{i,T}\Vert_{2}^{2}\Big]^{2}\\
			&\leq (\delta_{m}^{2}+G)^{2},
		\end{aligned}
\end{equation}
where step $(d)$ follows from $\mathbb{E}[X^{2}]{\leq}\mathbb{E}[X]^{2}$ and $\delta_{m}^{2}\triangleq \max\delta_{i}^{2}$. On the other hand, based on \eqref{avg2} the bound of $\mathbb{E}[\Vert \mathbf{w}_{T}{-}\tilde{\mathbf{w}}_{T} \Vert_{2}^{2}]$ is derived by
\begin{equation}
	\begin{aligned}
			\label{p19}
			\mathbb{E}[\Vert \mathbf{w}_{T} &- \tilde{\mathbf{w}}_{T} \Vert_{2}^{2}]= \mathbb{E}[\Vert \mathbf{P}\tilde{\mathbf{w}}_{T} - \tilde{\mathbf{w}}_{T} \Vert_{2}^{2}]\\
			&= \mathbb{E}[\Vert (\mathbf{P}-\mathbf{I}) \tilde{\mathbf{w}}_{T} \Vert_{2}^{2}]\leq \Vert \mathbf{P}-\mathbf{I} \Vert_{2}^{2}\mathbb{E}[\tilde{\mathbf{w}}_{T} \Vert_{2}^{2}].
		\end{aligned}
\end{equation}
Then, from \eqref{sgd2}, we have
	\begin{equation}
			\begin{aligned}
					\label{p110}
					\mathbb{E}&[\Vert\tilde{\mathbf{w}}_{T} \Vert_{2}^{2}]=\mathbb{E}[\Vert\mathbf{w}_{T-1}-\eta\mathbf{g}_{T-1} \Vert_{2}^{2}]\\
					&=\mathbb{E}[\Vert\mathbf{w}_{T-1}-\eta\mathbf{g}_{T-1}+\eta\nabla F_{T-1}-\eta\nabla F_{T-1} \Vert_{2}^{2}]\\
					&=\mathbb{E}[\Vert\mathbf{w}_{T-1}-\eta\nabla F_{T-1}+\eta(\nabla F_{T-1}-\mathbf{g}_{T-1}) \Vert_{2}^{2}]\\
					&\overset{(c)}{\leq}\mathbb{E}\Big[(1+\gamma_{1})\Vert\mathbf{w}_{T-1}-\eta\nabla F_{T-1}\Vert_{2}^{2}\\
					&~~~~~~~~+\Big(1+\frac{1}{\gamma_{1}}\Big)\eta^{2}\Vert\nabla F_{T-1}-\mathbf{g}_{T-1} \Vert_{2}^{2}\Big]\\
					&\leq (1+\gamma_{1})\Vert\mathbf{w}_{T-1}-\eta\nabla F_{T-1}\Vert_{2}^{2}+\Big(1+\frac{1}{\gamma_{1}}\Big)\eta^{2}\sigma,
			\end{aligned}
	\end{equation}
where
	\begin{equation}
	\begin{aligned}
		\label{p111}
		\Vert&\mathbf{w}_{T-1}-\eta\nabla F_{T-1}\Vert_{2}^{2}\\
		&=\mathbb{E}[\Vert\mathbf{w}_{T-1}-\eta\mathbf{g}_{T-1}+\eta\mathbf{g}_{T-1}-\eta\nabla F_{T-1}\Vert_{2}^{2}]\\
		&\overset{(c)}{\leq}\mathbb{E}\Big[(1+\gamma_{2})\Vert\mathbf{w}_{T-1}-\eta\mathbf{g}_{T-1}\Vert_{2}^{2}\\
		&~~~~~~~~+\Big(1+\frac{1}{\gamma_{2}}\Big)\eta^{2}\Vert\mathbf{g}_{T-1}-\nabla F_{T-1} \Vert_{2}^{2}\Big]\\
		&\leq (1+\gamma_{2})\mathbb{E}[\Vert\tilde{\mathbf{w}}_{T-1} \Vert_{2}^{2}]+\Big(1+\frac{1}{\gamma_{2}}\Big)\eta^{2}\sigma.
	\end{aligned}
\end{equation}
Let $\gamma_{1}=\gamma_{2}=(1+\delta_{m}\eta^{-\frac{1}{2}})^{(2T+1)^{-1}}-1$ and substituting \eqref{p111} into \eqref{p110}, the bound in \eqref{p110} can be written as
\begin{equation}
	\begin{aligned}
		\label{p112}
		\mathbb{E}&[\Vert\tilde{\mathbf{w}}_{T} \Vert_{2}^{2}]\leq (1+\gamma_{1})\Big( (1+\gamma_{2})\mathbb{E}[\Vert\tilde{\mathbf{w}}_{T-1} \Vert_{2}^{2}]\\
		&~~~~~~~~~~~~+\big(1+\frac{1}{\gamma_{2}}\big)\eta^{2}\sigma \Big)+\Big(1+\frac{1}{\gamma_{1}}\Big)\eta^{2}\sigma\\
		&=(1+\gamma_{1})^{2}\mathbb{E}[\Vert\tilde{\mathbf{w}}_{T-1} \Vert_{2}^{2}]+(3+\gamma_{1}+2\gamma_{1}^{-1})\eta^{2}\sigma\\
		&\overset{(e)}{\leq}\big((1+\gamma_{1})^{2T}-1\big)(\gamma_{1}+1)\gamma_{1}^{-2}\eta^{2}\sigma\\
		&\overset{(f)}{<}(1+\gamma_{1})^{2T+1}\eta^{2}\sigma\\
		&=\eta\sigma(\eta+\delta_{m}\sqrt{\eta}),
	\end{aligned}
\end{equation}
where step $(e)$ follows from $\tilde{\mathbf{w}}_{1}=\mathbf{0}$ when $\eta<1$, and step $(f)$ follows from $\gamma_{1}>1$. Thus, by substituting \eqref{p112} into \eqref{p19}, substituting the result and \eqref{p18} into \eqref{p17}, and let $\gamma=\eta$, we have
\begin{equation}
	\begin{aligned}
		\label{p113}
		\mathbb{E}&[F(\mathbf{w}_{T}){-}F(\tilde{\mathbf{w}}_{T})]\\
		&\leq \frac{\gamma}{2}(\delta_{m}^{2}{+}G)^{2}{+}\Big(\frac{1}{2\gamma}{+}\frac{M}{2}\Big)\Vert \mathbf{P}{-}\mathbf{I} \Vert_{2}^{2}\eta\sigma(\eta{+}\delta_{m}\sqrt{\eta})\\
		&= \frac{\eta}{2}(\delta_{m}^{2}{+}G)^{2}{+}\frac{\sigma}{2}(\eta{+}\delta_{m}\sqrt{\eta})\Vert \mathbf{P}{-}\mathbf{I} \Vert_{2}^{2}\\
		&~~~~+(\eta{+}\delta_{m}\sqrt{\eta})\frac{M\sigma\eta}{2}\Vert \mathbf{P}{-}\mathbf{I} \Vert_{2}^{2}\\
		&=\frac{\eta}{2}\big((\delta_{m}^{2}{+}G)^{2}{+}\sigma\big(1{+}(\eta{+}\delta_{m}\sqrt{\eta})M\big)\Vert \mathbf{P}{-}\mathbf{I} \Vert_{2}^{2}\big)\\
		&~~~~+\frac{\sqrt{\eta}}{2}\delta_{m}\Vert \mathbf{P}{-}\mathbf{I} \Vert_{2}^{2}\\
		&=\frac{\sqrt{\eta}}{2}\Big(\sqrt{\eta}\big((\delta_{m}^{2}{+}G)^{2}\\
		&~~~~+\sigma\big(1{+}(\eta{+}\delta_{m}\sqrt{\eta})M\big)\Vert \mathbf{P}{-}\mathbf{I} \Vert_{2}^{2}\big){+}\delta_{m}\Vert \mathbf{P}{-}\mathbf{I} \Vert_{2}^{2}\Big),
	\end{aligned}
\end{equation}
where
\begin{equation}
	\begin{aligned}
		\label{p114}
		\Vert \mathbf{P}{-}\mathbf{I} \Vert_{2}^{2}&=\frac{1}{N}\Big((N{-}1)^{2}\Vert\mathbf{I}_{k\times k}\Vert_{2}^{2}{+}N\Vert(1{-}N)\mathbf{I}_{k\times k}\Vert_{2}^{2}\Big)\\
		&=\frac{k}{N}(N-1)(2N-1)\triangleq\phi(N),
	\end{aligned}
\end{equation} 
since
 \[
\mathbf{P}-\mathbf{I} = \frac{1}{N}\begin{bmatrix}
	(1-N)\mathbf{I}_{k\times k} & \dots  & \mathbf{I}_{k\times k} \\
	\vdots & \ddots & \vdots \\
	\mathbf{I}_{k\times k}      & \dots   & (1-N)\mathbf{I}_{k\times k}
\end{bmatrix}_{kN \times kN.}  
\] 
Thus, let $\eta=\frac{1}{Mt^{4}}$, we have
\begin{equation}
	\begin{aligned}
		\label{p115}
		\sum\limits_{t=1}^{T}&\mathbb{E}[F_{i}(\mathbf{w}_{i,t}){-}F_{i}(\tilde{\mathbf{w}}_{i,t})]=\frac{1}{2\sqrt{M}}\Big(\sqrt{\eta}\big((\delta_{m}^{2}{+}G)^{2}\\
		&~~{+}\sigma\big(1{+}(\eta{+}\delta_{m}\sqrt{\eta})M\big)\phi(N)\big){+}\delta_{m}\phi(N)\Big){\cdot}\sum\limits_{t=1}^{T}\frac{1}{t^{2}}.
	\end{aligned}
\end{equation}
Furthermore, since $\lim\limits_{T\to+\infty}\sum_{t=1}^{T}\frac{1}{t^{2}}{=}\frac{\pi^{2}}{6}$, we have $\sum_{t=1}^{T}\frac{1}{t^{2}}<\frac{\pi^{2}}{6}$. By substituting it into \eqref{p115} we completes the proof.


\section{Proof of Corollary 1}
Note that, when $\delta_{m}=0$ and $\eta=\frac{1}{Mt^{4}}$, since $t>1$, we have
\begin{equation}
	\begin{aligned}
			\label{p31}
			h(\delta_{m},N,t)&{\leq}\frac{\pi^{2}}{12\sqrt{M}}\cdot\Big(G^{2}{+}\sigma\big(1{+}\frac{1}{M}\big)\phi(N)\Big)\frac{1}{\sqrt{M}t^{2}}\\
			&{\leq} \Big(\frac{\pi^{2}\big(G^{2}{+}\sigma\phi(N)\big)}{12M}{+}\frac{\sigma\phi(N)\pi^{2}}{12M^{2}}\Big)\frac{1}{t-1}.
		\end{aligned}
\end{equation}
In addition, let $\bar{\tau}_{i}=\frac{\sum_{t=1}^{T-1}\tau_{i,t}}{T-1}$, we have
\begin{equation}
	\begin{aligned}
		\label{p32}
	A_{1}(t)&= \frac{1}{N}\sum\limits_{i=1}^{N}(1{-}m\eta)^{\bar{\tau}_{i}(t-1)}\big(F_{i}(\mathbf{w}_{i,1}){-}F_{i}(\mathbf{w}_{i}^{*})\big)\\
	&\leq (1{-}m\eta)^{t-1}\cdot\frac{1}{N}\sum\limits_{i=1}^{N}\big(F_{i}(\mathbf{w}_{i,1}){-}F_{i}(\mathbf{w}_{i}^{*})\big)\\
	&\overset{(g)}{\leq}\frac{1}{1-(t-1)\ln(1-m\eta)}\cdot\big(F(\mathbf{w}_{1}){-}F(\mathbf{w}^{*})\big)\\
	&\leq \frac{F(\mathbf{w}_{1}){-}F(\mathbf{w}^{*})}{-\ln(1-m\eta)}\cdot\frac{1}{t-1},
	\end{aligned}
\end{equation}
where step $(g)$ follows from $a^{x}<\frac{1}{1-x\ln a}$, where $0<a<1$, and
\begin{equation}
	\begin{aligned}
		\label{p33}
		A_{2}(t,\tau_{i,t})&=\frac{1}{N}\sum\limits_{i=1}^{N}\frac{\sigma }{2mt^{4}}\Big(1+\sum\limits_{j=1}^{t-1}(1{-}m\eta)^{\sum\limits_{t=j}^{T-1}\tau_{i,j}}\Big)\\
		&\leq \frac{1}{N}\sum\limits_{i=1}^{N}\frac{\sigma }{2mt^{4}}\big(1+(t-1)\big)\\
		&\leq \frac{\sigma}{2m}\cdot \frac{1}{t-1}.
	\end{aligned}
\end{equation}
Thus, the convergence bound in \eqref{gap1} can be written as
\begin{equation}
	\begin{aligned}
		\label{p34}
		h(&\delta_{m},N,t)+A_{1}(t)+A_{2}(t,\tau_{i,t})\\
		&\leq\Big(\frac{\pi^{2}\big(G^{2}+\sigma\phi(N)\big)}{12M}+\frac{\sigma\phi(N)\pi^{2}}{12M^{2}}\\
		&
		+\frac{F(\mathbf{w}_{1}){-}F(\mathbf{w}^{*})}{-\ln(1-m\eta)}+\frac{\sigma}{2m}\Big)\cdot\frac{1}{t-1}=\mathcal{O}\Big(\frac{1}{t-1}\Big).
	\end{aligned}
\end{equation}
Moreover, since $\lim\limits_{t\to+\infty}\big(h(\delta_{m},N,t)+A_{1}(t)+A_{2}(t,\tau_{i,t})\big)=\lim\limits_{t\to+\infty}\mathcal{O}(\frac{1}{t-1})=0$, this rewritten bound is tight. Thus, $h(\delta_{m},N,t)+A_{1}(t)+A_{2}(t,\tau_{i,t})\leq \epsilon$ is equivalent to the rewritten bound in \eqref{p34} less than $\epsilon$. After some basic manipulations, we can arrive at \eqref{rate}. This completes the proof.


\section{Proof of Theorem 4}
Let $(1-\frac{m}{M})^{t}=a_{t}$, we have
\begin{equation}
	\begin{aligned}
		\label{p51}
		\sum\limits_{j=1}^{T}\Big(1{-}\frac{m}{M}\Big)^{\sum\limits_{t=j}^{T}\tau_{i,t}}=a_{T}{+}a_{T}a_{T-1}{+}\cdots{+}a_{T}a_{T-1}{\cdots} a_{1}.
	\end{aligned}
\end{equation}
Since $a_{t}\in (0,1]$, let $a_{T}=a_{T-1}=\cdots=a_{2}=1$, we have 
\begin{equation}
	\begin{aligned}
		\label{p52}
		a_{T}{+}a_{T}a_{T-1}{+}\cdots{+}a_{T}a_{T-1}{\cdots} a_{1}\leq T-1+a_{1}.
	\end{aligned}
\end{equation}
Thus, \eqref{P5a} is relaxed to $T-1+a_{1}\leq \zeta$, which is equal to 
\begin{equation}
	\begin{aligned}
		\label{p53}
		\tau_{i,1}\geq \frac{\ln(\zeta-T+1)}{\ln \big(1-\frac{m}{M}\big)}.
	\end{aligned}
\end{equation}
Similarly, let $a_{T}=a_{T-1}=\cdots=a_{3}=a_{1}=1$, \eqref{P5a} can be relaxed as $T-2+2a_{2}\leq \zeta$, which means
\begin{equation}
	\begin{aligned}
		\label{p54}
		\tau_{i,2}\geq \frac{\ln\frac{1}{2}(\zeta-T+2)}{\ln \big(1-\frac{m}{M}\big)}.
	\end{aligned}
\end{equation}
Repeating this manipulation from $t=1$ to $T$, we obtain that
\begin{equation}
	\begin{aligned}
		\label{p55}
		[\tau_{i,1},\cdots,\tau_{i,T}]\succeq \frac{1}{\ln \big(1{-}\frac{m}{M}\big)}\Big[\ln(\zeta{-}T{+}1),{\cdots},\ln\frac{1}{T}\zeta\Big].
	\end{aligned}
\end{equation}
Based on \eqref{p55} and applying the similar manipulations in Appendix C of \cite{dli}, we complete the proof.

\section{Proof of Lemma 1}
Based on Assumption 1, we have
\begin{equation}
	\begin{aligned}
		\label{p61}
		F(\mathbf{w})\geq F(\mathbf{w}'){+}\nabla F(\mathbf{w}')^{T}(\mathbf{w}{-}\mathbf{w}'){+}\frac{m}{2}\Vert \mathbf{w}{-}\mathbf{w}' \Vert_{2}^{2}, 
	\end{aligned}
\end{equation}
where the RHS of is a quadratic function with $\mathbf{w}$, when $\mathbf{w}'$ is fixed. Thus, its minimal value is obtained when $\mathbf{w}=\mathbf{w}'-\frac{1}{m}\nabla F(\mathbf{w}')$. Then we have
\begin{equation}
	\begin{aligned}
		\label{p62}
		F(\mathbf{w})&\geq F(\mathbf{w}'){+}\nabla F(\mathbf{w}')^{T}(\mathbf{w}{-}\mathbf{w}'){+}\frac{m}{2}\Vert \mathbf{w}{-}\mathbf{w}' \Vert_{2}^{2}\\
		&\geq  F(\mathbf{w}'){-}\frac{1}{m}\nabla F(\mathbf{w}')^{T}\nabla F(\mathbf{w}'){+}\frac{m}{2}\Big\Vert \frac{1}{m}\nabla F(\mathbf{w}') \Big\Vert_{2}^{2}\\
		&=F(\mathbf{w}'){-}\frac{1}{2m}\Vert\nabla F(\mathbf{w}')\Vert_{2}^{2}. 
	\end{aligned}
\end{equation}
After some basic manipulations, we can arrive at \eqref{L1}. This completes the proof.

\section{Proof of Lemma 2} 
From the Assumptions 2 and 3, we have
\begin{equation}
	\begin{aligned}
		\label{p71}
		\mathbb{E}[\Vert \mathbf{g}_{i,t}^{(j)} \Vert_{2}^{2}]&=\mathbb{E}[\Vert \mathbf{g}_{i,t}^{(j)}-\nabla {F}_{i,t}^{(j)} +\nabla {F}_{i,t}^{(j)} \Vert_{2}^{2}]\\
		&=\mathbb{E}[\Vert \mathbf{g}_{i,t}^{(j)}-\nabla {F}_{i,t}^{(j)}\Vert_{2}^{2}\\
		&~~+2(\mathbf{g}_{i,t}^{(j)}{-}\nabla {F}_{i,t}^{(j)})^{T}\nabla {F}_{i,t}^{(j)} +\Vert\nabla {F}_{i,t}^{(j)} \Vert_{2}^{2}]\\
		&=\mathbb{E}[\Vert \mathbf{g}_{i,t}^{(j)}-\nabla {F}_{i,t}^{(j)}\Vert_{2}^{2}]\\
		&~~+2(\mathbb{E}[\mathbf{g}_{i,t}^{(j)}]{-}\nabla {F}_{i,t}^{(j)})^{T}\nabla {F}_{i,t}^{(j)} +\Vert\nabla {F}_{i,t}^{(j)} \Vert_{2}^{2}\\
		&\leq \sigma+\Vert\nabla {F}_{i,t}^{(j)} \Vert_{2}^{2}.
	\end{aligned}
\end{equation}
This completes the proof.
}
 
\bibliographystyle{ieeetr}
\bibliography{ref1}

\vfill

\end{document}